\documentclass[final]{cvpr}

\usepackage{times}
\usepackage{epsfig}
\usepackage{graphicx}
\usepackage{amsmath}
\usepackage{amssymb}
\usepackage{enumerate}
\usepackage{comment}
\usepackage{color}
\usepackage{float}
\usepackage{subfigure}
\usepackage{multirow}
\usepackage{color}
\usepackage{booktabs}
\usepackage{tabularx}
\usepackage[misc]{ifsym}

\usepackage[pagebackref=false,breaklinks=true,colorlinks,bookmarks=false]{hyperref}

\begin{document}

%%%%%%%%% TITLE
\title{Region-aware Adaptive Instance Normalization for Image Harmonization}

\author{Jun Ling\textsuperscript{1}, Han Xue\textsuperscript{1}, Li Song\textsuperscript{1,2 \Letter}, Rong Xie\textsuperscript{1}, Xiao Gu\textsuperscript{1}\\ %\thanks{Corresponding author}
\textsuperscript{1}Institute of Image Communication and Network Engineering, Shanghai Jiao Tong University, China\\
% {\tt\small lingjun@sjtu.edu.cn}
% \and
% Li Song\textsuperscript{1,2}\\
\textsuperscript{2}MOE Key Lab of Artificial Intelligence, AI Institute, Shanghai Jiao Tong University, China\\
{\tt\small \{lingjun, xue\_han, song\_li, xierong, gugu97\}@sjtu.edu.cn}
}

\maketitle
\pagestyle{empty}
\thispagestyle{empty}

% This paper proposes adding a "region-aware" instance normalization step to improve image harmonization networks. The paper originally received two weak accepts and a weak reject. Post-rebuttal these reviewers were updated to one weak accept and a borderline with remarks about the state-of-the-art results demonstrated in this work as well as concerns about the novelty of the method. AC read the paper and agrees that while the idea is simple, it is effective and would be useful to the research community as a new benchmark for image harmonization performance. After discussion with the AC panel, the decision was to accept the paper.

%%%%%%%%% ABSTRACT
\begin{abstract}
  Image composition plays a common but important role in photo editing. To acquire photo-realistic composite images, one must adjust the appearance and visual style of the foreground to be compatible with the background. Existing deep learning methods for harmonizing composite images directly learn an image mapping network from the composite to the real one, without explicit exploration on visual style consistency between the background and the foreground images. To ensure the visual style consistency between the foreground and the background, in this paper, we treat image harmonization as a \textbf{style} transfer problem. In particular, we propose a simple yet effective Region-aware Adaptive Instance Normalization (RAIN) module, which explicitly formulates the visual \textbf{style} from the background and adaptively applies them to the foreground. With our settings, our RAIN module can be used as a drop-in module for existing image harmonization networks and is able to bring significant improvements. Extensive experiments on the existing image harmonization benchmark datasets show the superior capability of the proposed method. Code is available at \textcolor{magenta}{https://github.com/junleen/RainNet}. 
\end{abstract}

%%%%%%%%% BODY TEXT
\section{Introduction}
\label{introduction}

\begin{figure}
\begin{center}
   % \centering
   \includegraphics[width=1\linewidth]{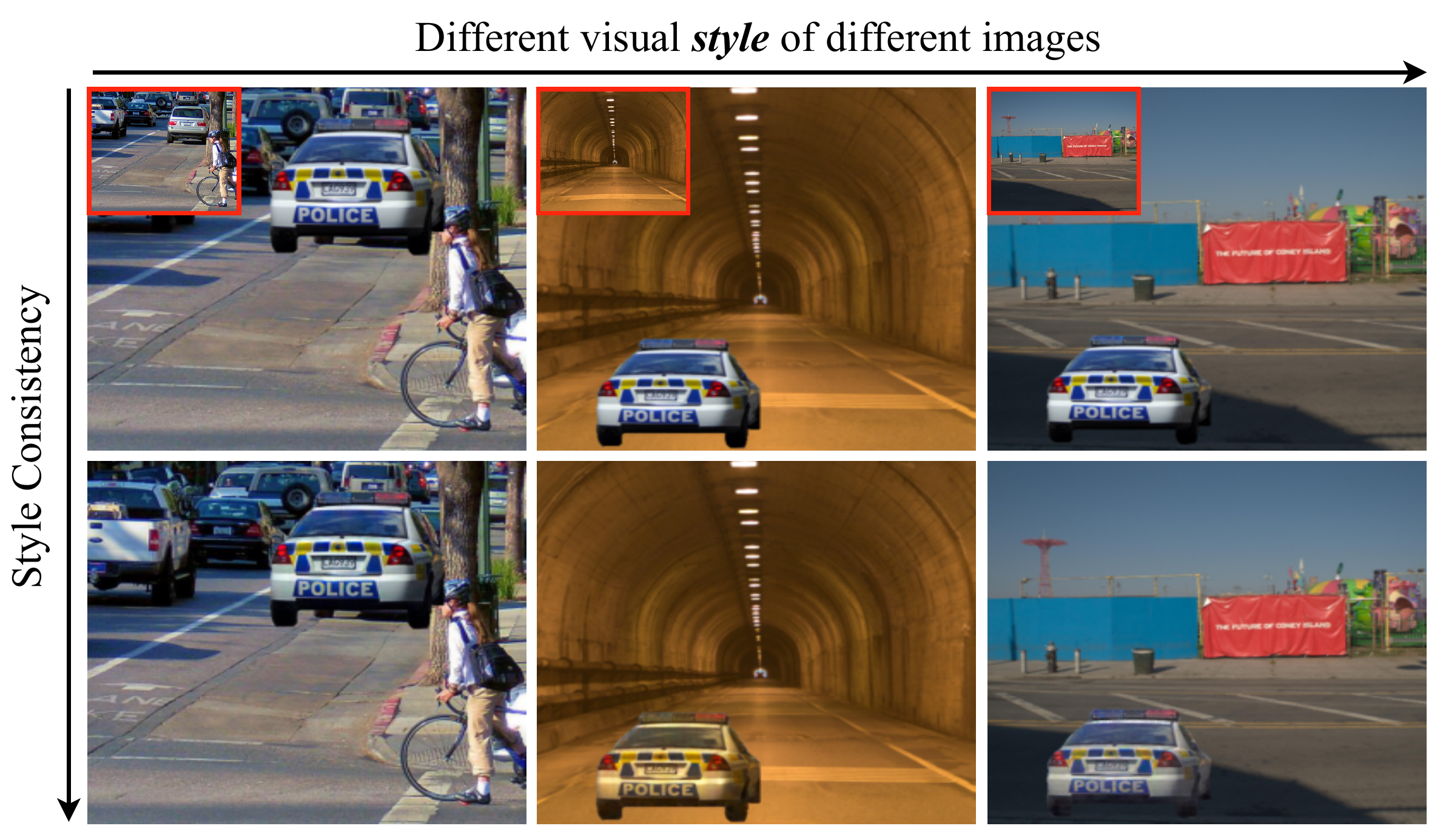}\\
\end{center}
   \caption{Illustration of our motivation. If we want to put a police car into these images with different visual \emph{style} , we must ensure that the car is compatible with the background images (small-sized images with red boundaries in the \emph{top row}). Simple cut-and-paste operations introduce unrealistic results (\emph{top row}). Our method aims to adaptively learn high-level visual \emph{style} from different backgrounds and produce harmonious composite images (\emph{bottom row}).}
\label{fig:toyexample}
\end{figure}

Image composition is one of the most common operations in image editing~\cite{xue2012understanding,cun2020improving} and data augmentation~\cite{dwibedi2017cut,zhang2020learning}, \etc. However, generating a realistic composite image by taking an object from one image and combining it with a new background image usually requires professional compositors to adjust the appearance of the foreground objects by photo editing software like Adobe Photoshop, and ensure the realism of the generated image. To alleviate this burden, image harmonization is introduced for adjusting the foreground and making it seamlessly integrated into the new image with less human involvement, especially for non-expert users.

However, what makes a composite image appear more realistic? In this paper, we present a new perspective for image harmonization. Let us take Fig.~\ref{fig:toyexample} for example. Fig.~\ref{fig:toyexample} shows three different real photos (small-sized images with red border) that hold different visual properties. When an unbefitting foreground object with special visual properties is pasted into a new image with incompatible visual features, we can easily distinguish it from real photos. This is an unsolved problem and has emerged for years, which we call visual \emph{style} discrepancy. Specifically, in this paper, we define the visual \emph{style} of an image as visual properties including illumination, color temperature, saturation, hue, texture \etc, which varies from image to image. To make a composite image look more realistic, we must ensure a more consistent visual \emph{style} between the foreground and the background.

Abundant image harmonization approaches have been proposed for improving the realism of composite images. Traditional methods address the harmonization problem by transferring statistics of hand-crafted features between foreground and background regions, such as color~\cite{pitie2007linear,reinhard2001color,xue2012understanding,sunkavalli2010multi}. However, these methods only work in simple cases where the foreground image is already consistent with the background image. Recently, more deep learning-based methods~\cite{cong2020dovenet,cun2020improving,tsai2017deep,zhu2015learning} have been proposed for generating harmonious images in an end-to-end manner. Zhu \etal~\cite{zhu2015learning} propose to adopt a discriminative model to predict the realism of a compsite image and assist optimization of color adjustment. Tsai \etal~\cite{tsai2017deep} propose an end-to-end learning approach for image harmonization while only constraining semantic information learning in the encoder. Cun \etal~\cite{cun2020improving} adopt a spatial-separated attention module to enforce the network to learn the foreground and background features separately, failing to ensure the \emph{style} consistency between these two parts. To sum up, none of these methods really consider the realism from the perspective of visual \emph{style} consistency. Cong \etal~\cite{cong2020dovenet} propose to use a domain verification discriminator and adversarial loss~\cite{goodfellow2014generative} to improve domain-consistency between foreground and background regions but neglect to explicitly transform the foreground features in the generator. However, performance improvement brought by such an auxiliary discriminator is limited (\ie, 0.27dB for PSNR, which is revealed in~\cite{cong2020dovenet}).

To address these issues, in this work, we reframe image harmonization as a background-to-foreground \emph{style} transfer problem, where we render the foreground image to hold similar visual \emph{style} of the background image. Taking \emph{style} guidance from background information is of great importance because the foreground image should be converted to own different appearances when pasted into different background images (as illustrated in Fig.~\ref{fig:toyexample}). To generate style-consistent and realistic-looking composite images, we expect a unified transferring operation to adaptively adjust the \emph{style} of the foreground objects to be in perfect harmony with new background images even collected in different environments. Therefore, in this work, we propose a learnable layer, named Region-aware Adaptive Instance Normalization (RAIN) layer, to learn the style from background images and apply it to the foreground objects. By taking convolutional features and the foreground mask as input, the RAIN layer aligns the channel-wise mean and variance of the foreground activation to match those learned from the background. The details of the proposed RAIN module are presented in Fig.~\ref{fig:rain}. It is worth mentioning that our RAIN layer can be easily applied to existing image harmonization networks and encourage performance improvements. 

The contributions of this work are as follows. 
% \begin{enumerate}
% \noindent
% \item[$\bullet$] 
1) To the best of our knowledge, we are the first to introduce the \emph{style} concept of background images and regard the image harmonization task as a \emph{style} transferring problem. 
% \noindent
% \item[$\bullet$] 
2) We propose a novel Region-aware Adaptive Instance Normalization (RAIN) method, which captures the \emph{style} information only from the background features and applies it to the foreground for image harmonization tasks. Our RAIN module is simple yet effective and can be used as a \emph{plug-and-play} module for existing image harmonization networks to enhance their performance. 
% \noindent
% \item[$\bullet$] 
3) Extensive experiments demonstrate that our method surpasses the state-of-the-art methods by a large margin. 
% \end{enumerate}

\begin{figure*}[t]
\begin{center}
   % \centering
   \includegraphics[width=0.9\linewidth]{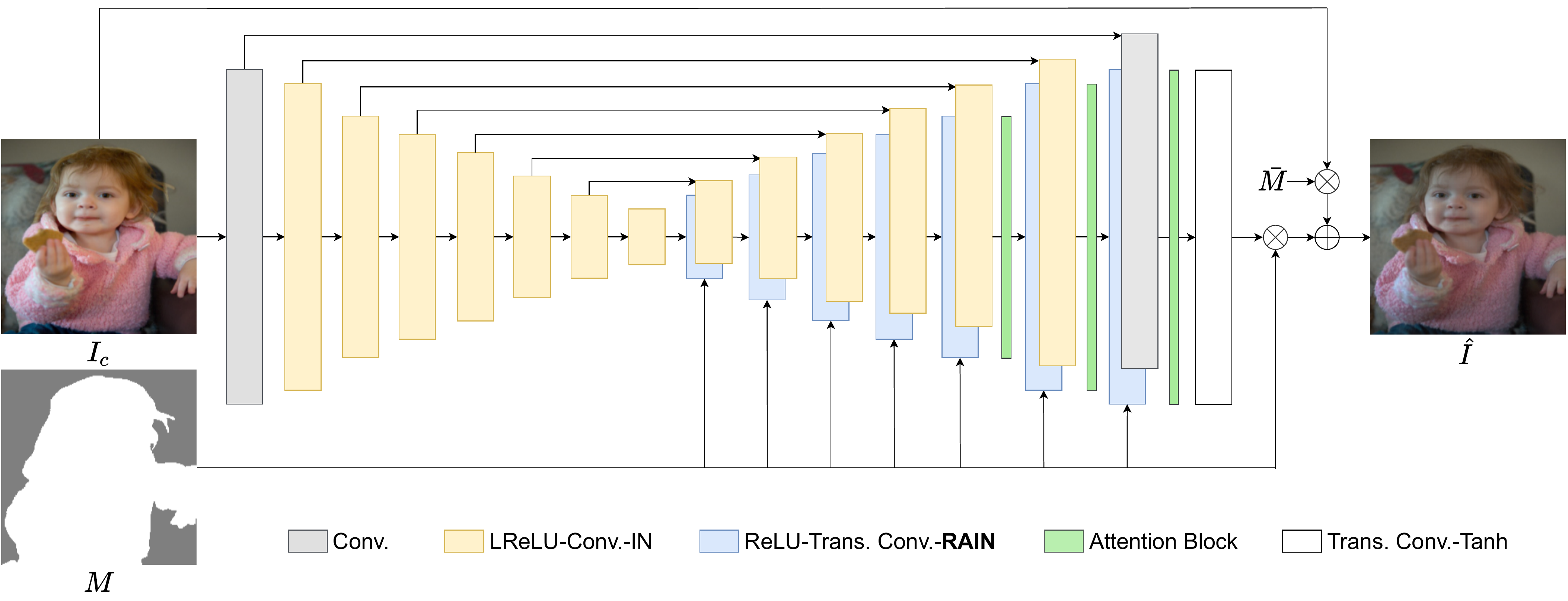}
\end{center}
   \caption{Overview of the proposed generator. We provide a detailed structure of our RainNet to ensure better understanding and reproducibility. The bottom legend: Conv.= Convolution, Trans. = Transposed. }
\label{fig:model_attentioned}
\end{figure*}

%-------------------------------------------------------------------------
\section{Related work}

\noindent
\textbf{Image harmonization} aims to adjust a foreground image to seamlessly match a background image. Traditional methods mainly focus on matching the appearance of the foreground with background regions based on handful of hand-crafted heuristics, such as color statistics~\cite{reinhard2001color,pitie2007linear,xue2012understanding}, gradient information~\cite{jia2006drag,perez2003poisson,tao2013error}, multi-scale statistical features~\cite{sunkavalli2010multi}, semantic information~\cite{tsai2017deep,tsai2016sky}. These methods directly match appearance to harmonize a composite image while paying less attention to visual realism. Johnson \etal~\cite{johnson2010cg2real} introduce a data-driven approach to improve the realism of computer-generated images by retrieving a small number of real images from an image dataset and transfer the features of color, tone, texture, \etc. Lalonde \etal~\cite{lalonde2007using} predict the realism of images by learning global and local statistics from natural images. With the advances of deep learning, more deep learning-based methods~\cite{cun2020improving,cong2020dovenet,tsai2017deep,zhu2015learning} draw much attention due to their impressive results. % Zhu \etal~\cite{zhu2015learning} train a discriminative model to predict the realism of composite images and then adjust the color of the foreground region with supervision from the discriminative model. Tsai \etal~\cite{tsai2017deep} propose to perform image harmonization with an end-to-end encoder-decoder network and introduce an auxiliary segmentation head to learn semantic representation. Cun \etal~\cite{cun2020improving} develop an attentive module to learn features in the foreground and background regions individually. Cong \etal~\cite{cong2020dovenet} predicts the realism of composition from the perspective of image domain. 
Different from these works, we start from the perspective of background-to-foreground \emph{style} transfer, and push the limit of image harmonization performance by introducing a novel RAIN module, which separates our approach from previous methods.

\medskip
\noindent
\textbf{Neural style transfer} is designed to render a photo with special visual style captured from artistic creations while retaining the content information from the original image. Earlier style transfer methods concentrate on texture synthesis or transfer~\cite{efros2001image,elad2017style,li2016precomputed,ulyanov2016texture}. Gatys \etal~\cite{gatys2016image} first introduce a method to match feature statistics in pre-trained convolutional networks and demonstrate impressive artistic style transfer. To achieve the goal of real-time style transfer, Johnson \etal~\cite{johnson2016perceptual} propose a novel feed-forward perceptual loss with a pre-trained VGG network~\cite{simonyan2014very}. Later, Huang \etal~\cite{huang2017arbitrary} propose Adaptive Instance Normalization (AdaIN) to achieve arbitrary style transfer from the perspective of feature normalization. Besides AdaIN, other normalization methods~\cite{dumoulin2016learned,ulyanov2016instance} were also proposed for fast stylization and later adopted in various vision tasks~\cite{huang2018multimodal,li2020advancing,liu2019few,zakharov2019few,xue2020realistic}. 

\medskip
\noindent
\textbf{Normalization layers} include unconditional normalization (Batch Normalization (BN)~\cite{ioffe2015batch}, Instance Normalization (IN)~\cite{ulyanov2016instance}, Layer Normalization (LN)~\cite{ba2016layer}, Group Normalization (GN)~\cite{wu2018group}, \etc) and conditional normalization (Conditional Batch Normalization (CBN)~\cite{de2017modulating}, Conditional Instance Normalization (CIN)~\cite{dumoulin2016learned}, SPADE~\cite{park2019semantic}, Region Normalization (RN)~\cite{yu2020region}, and AdaIN~\cite{huang2017arbitrary}, \etc). Note that unconditional normalization aligns the mean and variance of feaures without guidance from external data. On the contrary, conditional normalization~\cite{de2017modulating,dumoulin2016learned,huang2017arbitrary,park2019semantic} requires external data to provide affine parameters, which embed new information from the external data. SPADE~\cite{park2019semantic} applies spatially-varying transformations from semantic masks for image synthesis, which cannot be used in our image harmonization task due to the irregular shapes of foreground objects. RN~\cite{yu2020region} is designed for image inpainting which aims to alleviate the mean and variance shift problem but it does not consider the semantic connection between the background and the foreground. AdaIN~\cite{huang2017arbitrary} is proposed for real-time image stylization which uses a pre-trained VGG network to extract style code. However, it is not practical for our task because the \emph{style} defined in this work is considered to be consistent with image realism instead of texture. Besides, the background image with one region removed cannot be extracted by a pre-trained network, which will introduce new problems of mean and variance shift. In this paper, we seek ways to establish a connection between the background and the foreground. Therefore, we regard image harmonization as a new \emph{style} transfer task in which we transfer \emph{style} from the background to the foreground instance. 

%------------------------------------------------------------------------
\section{Our approach}
\label{sec:approach}
Our goal is to learn a mapping network for the foreground image and ensure that the foreground image is compatible with the background. To achieve this goal, we introduce our Region-aware Adaptive Instance Normalization (RAIN) for improving the performance of basic networks. 

%-------------------------------------------------------------------------
\subsection{Problem formulation}
\label{subsec:problem_formulation}
We consider a foreground image and a background image as $I_{f}$ and $I_{b}$ respectively. The foreground mask is denoted by $M$, which indicates the region to be harmonized in the composite image $I_{c}$. Accordingly, the background mask is $\bar{M}=1-M$. The object composition process is formulated as $I_{c} = M\circ I_{f} + (1-M) \circ I_{b}$, 

where $\circ$ is the Hadamard product. In this paper, we define the harmonization model as generator $G$, and the harmonized image as $\hat{I} = G(I_{c}, M)$, where $G$ is a learnable model that we expect to optimize for making $\hat{I}$ close to the ground truth image $I$ by $\|G(I_{c}, M) - I \|_{1}$.

%-------------------------------------------------------------------------

\subsection{Region-aware Adaptive Instance Normalization (RAIN)}
\label{subsec:RAIN}
\begin{figure}
\begin{center}
   % \centering
   \includegraphics[width=1\linewidth]{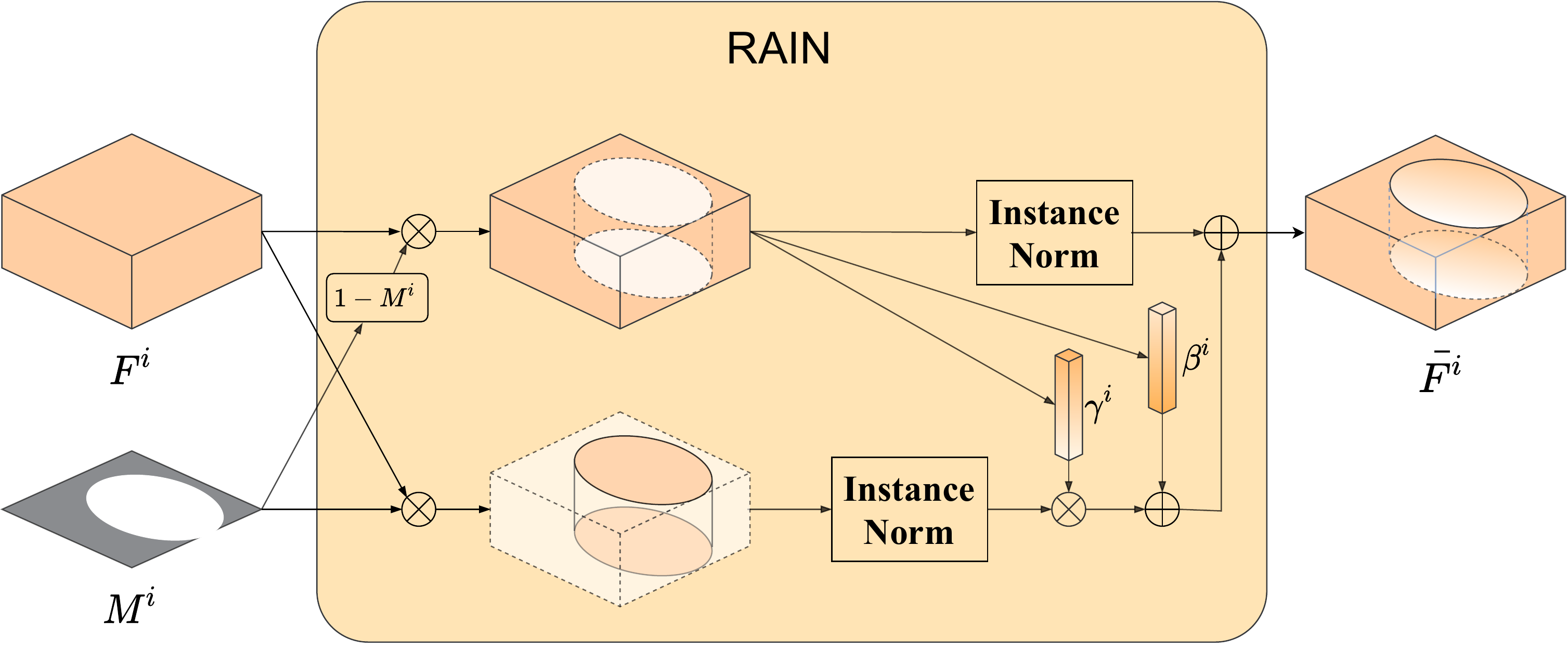}
\end{center}
   \caption{Our RAIN module takes the input feature $F^i$ and resized mask $M^i$ as input. Then we obtain the statistical style parameters $\gamma^i$ and $\beta^i$ from only background features. The produced $\gamma^i$ and $\beta^i$ are multiplied and added to the normalized foreground features in a channel-wise manner. %The background features are normalized by IN. 
   }
   \label{fig:rain}
\end{figure}

The input of our normalization module consists of two parts, \ie, the foreground mask, and the convolutional features (see in Fig.~\ref{fig:rain}). Without loss of generality, we take the RAIN module in the $i$-th layer of $G$ for example. Let $F^{i}\in \mathbb{R}^{H^{i}\times W^{i}\times C^{i}}$ be the activations and $M^{i}\in \mathbb{R}^{H^{i}\times W^{i}}$ be the resized foreground mask in the $i$-th layer, where $H^{i}, W^{i}, C^{i}$ denote the height, width, and number of channels of feature $F^{i}$, respectively. We propose a simple yet effective normalizing method called Region-aware Adaptive Instance Normalization (RAIN).

As depicted in Fig.~\ref{fig:rain}, we first multiply the input features $F^{i}$ by the foreground mask and its corresponding background mask. Then we normalize the foreground features by IN~\cite{ulyanov2016instance}, and then affine the normalized features with learned scale and bias from the background features. The new activation value $\bar{F^{i}}$ at site ($h, w, c$) in the foreground region is computed by:
\begin{equation}
\label{equ:rain}
\bar{F^{i}}_{h, w, c}=\gamma_{c}^{i}\frac{F^{i}_{h,w,c}-\mu_{c}^{i}}{\sigma_{c}^{i}} + \beta_{c}^{i},
\end{equation}
where $\mu_{c}^{i}$ and $\sigma_{c}^{i}$ are the channel-wise mean and variance of the foreground feature in $i$-th layer:
\begin{equation}
\label{equ:mean_foreground}
\mu_{c}^{i} = \frac{1}{\#\{M^{i}=1\}}\sum_{h,w}F^{i}_{h,w,c}\circ M^{i}_{h,w},
\end{equation}
\begin{equation}
\label{equ:var_foreground}
\sigma_{c}^{i} = \sqrt{\frac{1}{\#\{M^{i}=1\}}\sum_{h,w}(F^{i}_{h,w,c}\circ M^{i}_{h,w}-\mu_{c}^{i})^2 + \epsilon} .
\end{equation}
The expression $\#\{x=k\}$ means the number of pixels which equal to value $k$ in $x$. The $\gamma_{c}^{i}$ and $\beta_{c}^{i}$ are the mean and standard deviation of the activations of the background in channel $c$ of layer $i$:
\begin{equation}
\label{equ:mean_background}
\gamma_{c}^{i} = \frac{1}{\#\{\bar{M}^{i}=1\}}\sum_{h,w}F^{i}_{h,w,c}\circ \bar{M}^{i}_{h,w}
\end{equation}
\begin{equation}
\label{equ:var_background}
\beta_{c}^{i} = \sqrt{\frac{1}{\#\{\bar{M}^{i}=1\}}\sum_{h,w}(F^{i}_{h,w,c}\circ \bar{M}^{i}_{h,w}-\gamma_{c}^{i})^2 + \epsilon}
\end{equation}
where $\bar{M}^{i}$ is the background mask in $i$-th layer.

Our method is different from AdaIN in two aspects. First, our method focuses on transferring the visual \emph{style} from background to foreground only within the same image while AdaIN considers the style of features from another whole external image. Second, AdaIN uses a pre-trained VGG network to extract and calculate the statistics of the features, which cannot be directly employed in our task. Contrarily, our RAIN is designed and trained for image harmonization, such that the style parameters are better fitted for the foreground adjustment operations. Moreover, comprehensive experimental results demonstrate the efficacy of the proposed method. %simple yet effective 

\noindent
\textbf{RainNet.} We take a simple U-Net~\cite{ronneberger2015u,isola2017image} alike network without any feature normalization layers as our basic network architecture. Following~\cite{cong2020dovenet,cun2020improving}, in this work, we add three attention blocks in the decoder part for our \textbf{Baseline} network. Theoretically, our RAIN module can be applied in any layers of the basic network. In this work, we train our baseline with different normalization methods and exploit the design strategy of implementing our RAIN module to obtain the best model, denoted as RainNet. The structure of our RainNet is depicted in Fig.~\ref{fig:model_attentioned}.

\noindent
\textbf{Why is RAIN effective?} Briefly, RAIN helps the model to capture the visual \emph{style} information from the background image and inject it into the foreground, so that the generated foreground objects are more compatible with the new background. 

Consider a simple case with Region Normalization (RN)~\cite{yu2020region} that performs feature normalization for the foreground features and the background features separately. In each normalization layer, the background features will not provide any guidance for the model to transform the foreground features. Consequently, the model can only transform the foreground image to hold the average back-ground visual statistics in the training data, leading to unsatisfactory harmonizing results. However, when performing normalization with BN or IN, the foreground features will be normalized with the same mean and variance as the background features, where the mean and variance are statistically measured from the whole global feature map. Unfortunately, the styles of background features will be shifted by those statistics from the foreground and limit the style consistency learning in subsequent layers. 

In contrast with other normalization methods, our RAIN module only transfers the statistics from the background features to the normalized foreground features, without the influences from inconsistent foreground objects. As plotted in Fig.~\ref{fig:with_normalization}, IN and BN outperform RN, while our RAIN outperforms IN and BN by a large margin, demonstrating the reasonableness of our aforementioned analysis.

\begin{table*}[!htp]
\small
\begin{center}
\begin{tabular}{rcccccc}
\toprule
{Method} & {Venue} & {HCOCO} & {HAdobe5k} & {HFlickr} & {Hday2night} & {Average} \\
\midrule
Input composite & - & 33.94 & 28.16 & 28.32 & 34.01 & 31.63 \\
% \hline
Lalonde and Efros~\cite{lalonde2007using} & ICCV'07 & 31.14 & 29.66 & 26.43 & 29.80 & 30.16 \\
% \hline
Xue \etal~\cite{xue2012understanding} & TOG'12 & 33.32 & 28.79 & 28.32 & 31.24 & 31.40 \\
% \hline
Zhu \etal~\cite{zhu2015learning} & ICCV'15 & 33.04 & 27.26 &  27.52 & 32.32 & 30.72 \\
% \hline
DIH~\cite{tsai2017deep} & CVPR'17 & 34.69 & 32.28 & 29.55 &  34.62 & 33.41 \\
% \hline
S$^2$AM~\cite{cun2020improving} & TIP'20 & 35.47 & 33.77 &  30.03 & 34.50 & 34.35 \\
% \hline
DoveNet~\cite{cong2020dovenet} & CVPR'20 & \textcolor{blue}{\underline{35.83}} & \textcolor{blue}{\underline{34.34}} & \textcolor{blue}{\underline{30.21}} & \textcolor{red}{\textbf{35.18}} & \textcolor{blue}{\underline{34.75}} \\
\midrule
Baseline & This work & 35.03 & 33.35 & 29.50 & \textcolor{blue}{\underline{35.02}} & 33.92 \\
RainNet & Ours  & \textcolor{red}{\textbf{37.08}} & \textcolor{red}{\textbf{36.22}} & \textcolor{red}{\textbf{31.64}} & 34.83 & \textcolor{red}{\textbf{36.12}} \\
\bottomrule
\end{tabular}
\end{center}
\caption{Quantitative performance comparisons of PSNR metric on the four sub-datasets of iHarmoni4~\cite{cong2020dovenet}. The numbers in \textcolor{red}{\textbf{red}} and \textcolor{blue}{\underline{blue}} represent the best and second best performance. As can be found from the results, our approach performs favorably against other methods. }
\label{tab:performance_on_4datasets}
\end{table*}

\begin{table*}[!htp]
\small
\begin{center}
% \centering
\begin{tabular}{rccccccccc}
\toprule
\multirow{2}{*}{Method} & \multirow{2}{*}{Venue} & \multicolumn{2}{c}{0$\%\sim$5$\%$} & \multicolumn{2}{c}{5$\%\sim$15$\%$} & \multicolumn{2}{c}{15$\%\sim$100$\%$} & \multicolumn{2}{c}{Average} \\
\cmidrule(r){3-4}
\cmidrule(r){5-6}
\cmidrule(r){7-8}
\cmidrule(r){9-10}
& & MSE  & fMSE  & MSE  & fMSE  & MSE  & fMSE  & MSE  & fMSE  \\
\midrule
% Input composite & - & 28.51 & 1208.86 & 119.19 & 1323.23 & 577.58 & 1887.05 & 172.47 & 1387.30  \\
% \hline
Lalonde and Efros~\cite{lalonde2007using} & ICCV'07 & 41.52 & 1481.59 & 120.62 & 1309.79 & 444.65 & 1467.98 & 150.53 & 1433.21  \\
% \hline
Xue \etal~\cite{xue2012understanding} & TOG'12 & 31.24 & 1325.96 & 132.12 & 1459.28 & 479.53 & 1555.69 & 155.87 & 1141.40  \\
% \hline
Zhu \etal~\cite{zhu2015learning} & ICCV'15 & 33.30 & 1297.65 & 145.14 & 1577.70 & 682.69 & 2251.76 & 204.77 & 1580.17\\
% \hline
DIH~\cite{tsai2017deep} & CVPR'17 & 18.92 & 799.17 & 64.23 & 725.86 & 228.86 & 768.89 & 76.77 & 773.18  \\
% \hline
S$^2$AM~\cite{cun2020improving} & TIP'20 & 15.09 & 623.11 & 48.33 & 540.54 & 177.62 & 592.83 & 59.67 & 594.67 \\
% \hline
DoveNet~\cite{cong2020dovenet} & CVPR'20 & \textcolor{blue}{\underline{14.03}} & \textcolor{blue}{\underline{591.88}} & \textcolor{blue}{\underline{44.90}} & \textcolor{blue}{\underline{504.42}} & \textcolor{blue}{\underline{152.07}} & \textcolor{blue}{\underline{505.82}} & \textcolor{blue}{\underline{52.36}} & \textcolor{blue}{\underline{549.96}}  \\
\midrule
Baseline & This work & 19.21 & 841.61 & 64.54 & 749.36 & 241.15 & 803.05 & 79.97 & 808.68 \\
RainNet & Ours & \textcolor{red}{\textbf{11.66}} & \textcolor{red}{\textbf{550.38}} & \textcolor{red}{\textbf{32.05}} & \textcolor{red}{\textbf{378.69}} & \textcolor{red}{\textbf{117.41}} & \textcolor{red}{\textbf{389.80}} & \textcolor{red}{\textbf{40.29}} & \textcolor{red}{\textbf{469.60}} \\
\bottomrule
\end{tabular}
\end{center}
\caption{We measure the error of different methods in foreground ratio range based on the whole test set. fMSE indicates the mean square error of the foreground region. The numbers in \textcolor{red}{\textbf{red}} and \textcolor{blue}{\underline{blue}} indicate the best and second-best results. }
\label{tab:performance_on_foreground_ratio}
\end{table*}

%------------------------------------------------------------------------
\section{Implementation}
\label{sec:implementation}

\noindent
\textbf{Datasets.} To demonstrate the efficacy of our approach, we analyze the performance of our model against previous methods on the benchmark dataset iHarmony4~\cite{cong2020dovenet}. According to~\cite{cong2020dovenet}, iHarmony4 consists of 4 sub-datasets (\ie, HCOCO, HAdobe5K, HFlicker and Hday2night), and 73147 pairs of synthesized composite images and corresponding ground truth images are provided. In our experiments, we follow the train-test split as~\cite{cong2020dovenet} suggested. % where 65742 pairs for training and 7404 pairs for testing, and present experimental results in Section~\ref{sec:results}. 

\noindent
\textbf{Training.} We trained the model by Adam~\cite{kingma2014adam} optimizer with a learning rate of 0.0002, and optimized our model with the same objective that DoveNet~\cite{cong2020dovenet} uses. Our model was optimized for 100 epochs on an Nvidia GTX 2080Ti GPU, with input images resized to 256$\times$256 and batch size set to 12. Detailed training objectives of our model are presented in the supplementary materials. %All experiments were conducted in PyTorch~\cite{paszke2017automatic} environment. 
% and ($\lambda_1$, $\lambda_2$, $\lambda_{3}$) are set to (1, 1, 100)

% -----------------------------------------------------------------------------------------------

\begin{figure*}[!htbp]
\begin{center}
   % \centering
   \includegraphics[width=1\linewidth]{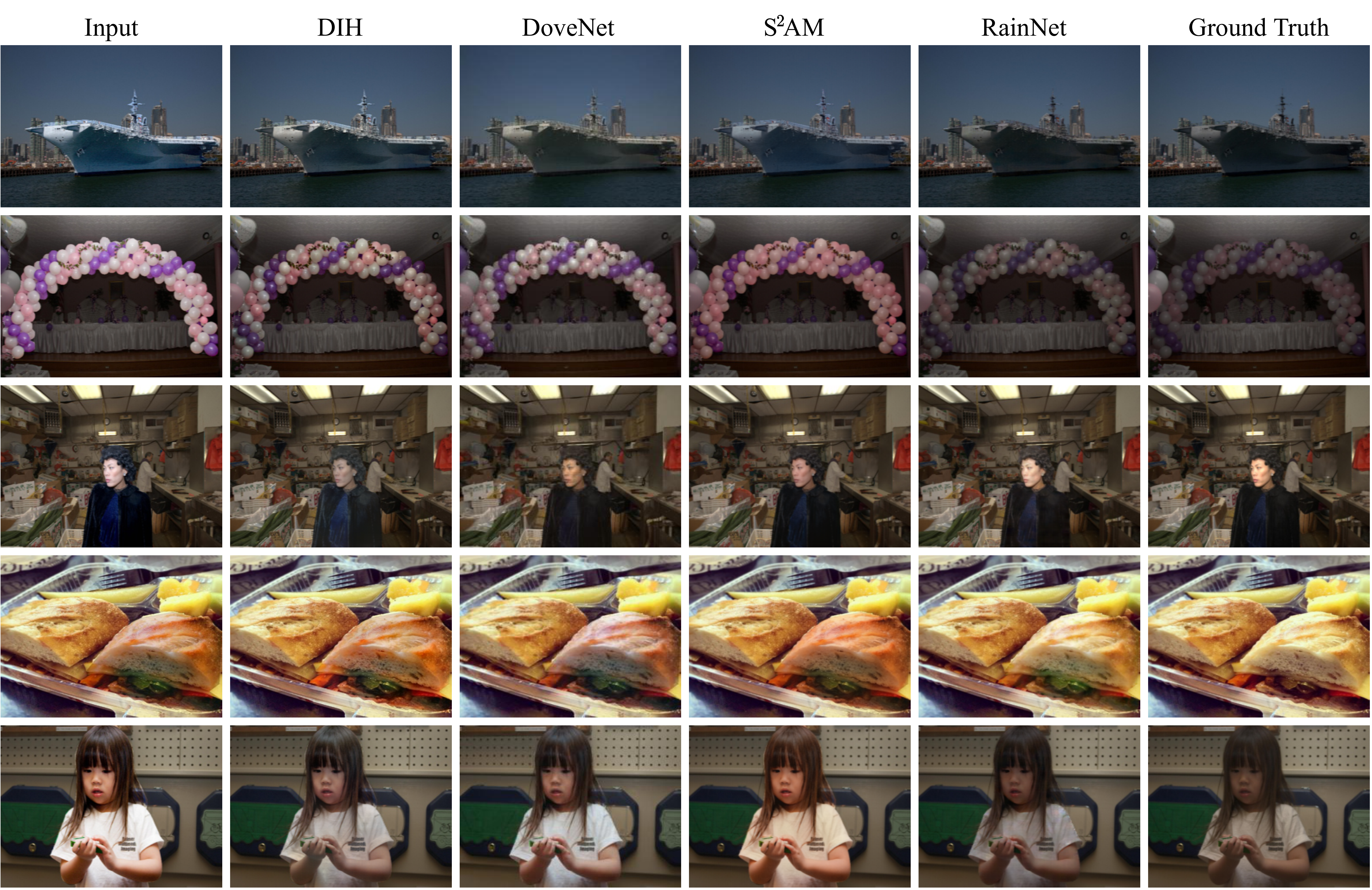}
\end{center}
   \caption{\textbf{Qualitative comparison}. We present example results of our RainNet against three state-of-the-art methods. The samples are taken from the testing dataset of iHarmony4~\cite{cong2020dovenet}. }
   \label{fig:comparison1}
\end{figure*}

\section{Experimental Results}
\label{sec:results}
In this section, we conduct extensive experiments to demonstrate the efficacy of our method. We first compare our best model (RainNet) to current state-of-the-art methods both qualitatively and quantitatively in Sec.~\ref{subsec:comparison}. Then, we investigate the design choice of RAIN for our generator in Sec.~\ref{subsec:ablation}. Subjective evaluations and further discussions are presented in Sec.~\ref{subsec:user_study} and Sec.~\ref{subsec:limitation_discussion}, respectively.

\begin{figure*}[!htbp]
\begin{center}
   % \centering
   \includegraphics[width=1\linewidth]{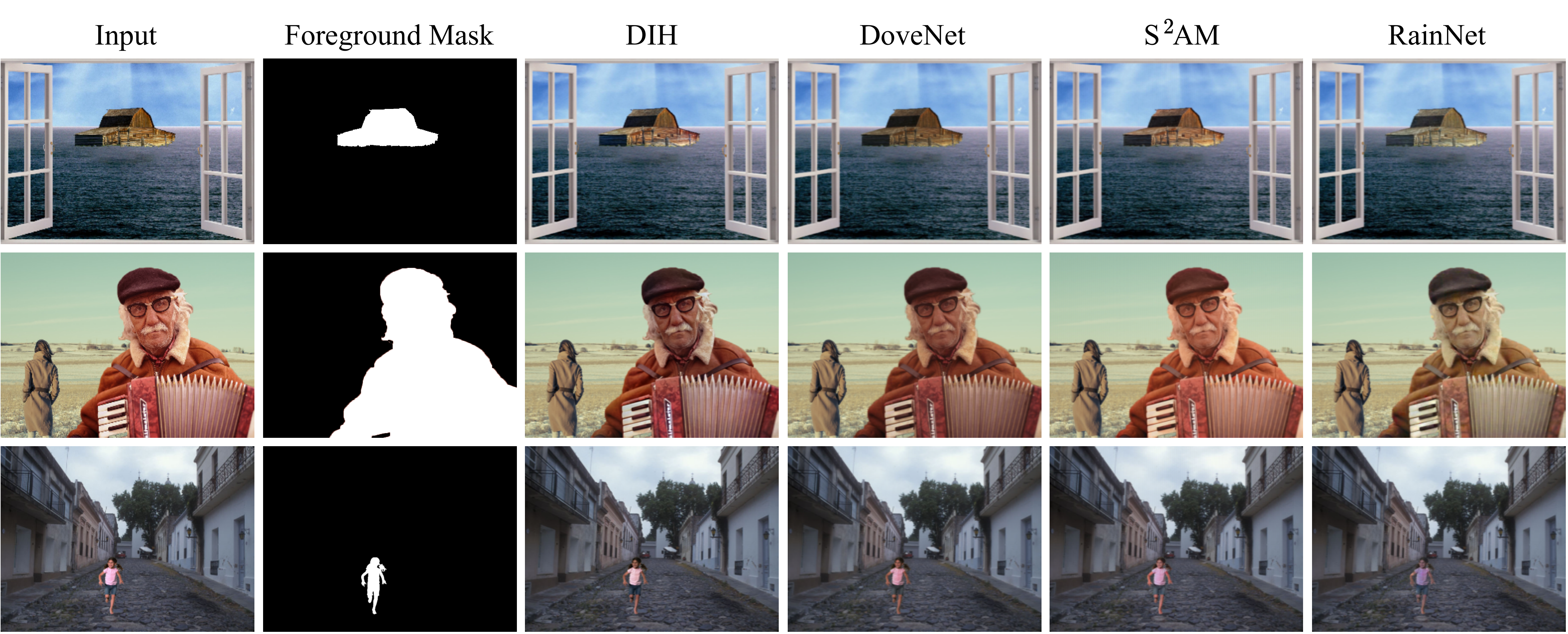}
\end{center}
   \caption{\textbf{Example results on real composite images.}. We present real composite images, foreground mask, the results of three state-of-the-art methods, and the proposed model. The samples are taken from the testing dataset of~\cite{tsai2017deep}. Our method achieves better harmonized visual results than competing methods. }
   \label{fig:comparison2}
\end{figure*}

\begin{table*}[!htp]
\footnotesize
\begin{center}
% \centering
\begin{tabular}{rp{0.5cm}<{\centering}p{0.6cm}<{\centering}p{0.6cm}<{\centering}p{0.5cm}<{\centering}p{0.6cm}<{\centering}p{0.6cm}<{\centering}p{0.6cm}<{\centering}p{0.6cm}<{\centering}p{0.6cm}<{\centering}p{0.5cm}<{\centering}p{0.6cm}<{\centering}p{0.6cm}<{\centering}p{0.6cm}<{\centering}p{0.6cm}<{\centering}p{0.6cm}<{\centering}}
% {lccccccccccccccccc}
\toprule
\multirow{2}{*}{Method} & \multicolumn{3}{c}{0$\%\sim$5$\%$} & \multicolumn{3}{c}{5$\%\sim$15$\%$} & \multicolumn{3}{c}{15$\%\sim$30$\%$} & \multicolumn{3}{c}{30$\%\sim$100$\%$}& \multicolumn{3}{c}{Average} \\
\cmidrule(r){2-4}
\cmidrule(r){5-7}
\cmidrule(r){8-10}
\cmidrule(r){11-13}
\cmidrule(r){14-16}
 & fL1  & PSNR & SSIM  & fL1  & PSNR & SSIM &  fL1  & PSNR & SSIM &  fL1  & PSNR & SSIM &  fL1  & PSNR & SSIM \\
\midrule
Baseline & 21.76 & 37.99 & 0.9951 & 20.55 & 32.05 & 0.9838 & 20.97 & 27.85 & 0.9631 & 21.49 & 24.39 & 0.9285 & 21.31 & 33.92 & 0.9824 \\
+ IN~\cite{ulyanov2016instance} & 18.61 & 39.08 & 0.9959 & 16.53 & 33.75 & 0.9870 & 16.34 & 29.77 & 0.9711 & 17.97 & 25.97 & 0.9384 & 17.69 & 35.32 & 0.9855 \\
+ BN~\cite{ioffe2015batch} & 17.81 & 39.48 & 0.9962 & 16.79 & 33.60 & 0.9876 & 17.76 & 29.15 & 0.9704 & 19.32. & 25.10 & 0.9395 & 17.65 & 35.34 & 0.9859 \\
+ RN~\cite{yu2020region} & 18.85 & 38.74 & 0.9959 & 17.54 & 32.85 & 0.9864 & 18.77 & 28.42 & 0.9673 & 20.55 & 24.37 & 0.9326 & 18.62 & 34.57 & 0.9842 \\
\midrule 
% \midrule
+ RAIN-1 & \textcolor{red}{\textbf{17.10}} & \textcolor{red}{\textbf{39.67}} & \textcolor{red}{\textbf{0.9963}} & \textcolor{blue}{\underline{14.70}} & \textcolor{blue}{\underline{34.69}} & \textcolor{blue}{\underline{0.9882}} & 14.20 & 31.02 & {0.9742} & 14.92 & 27.36 & 0.9478 & \textcolor{red}{\textbf{15.88}} & \textcolor{blue}{\underline{36.06}} & \textcolor{blue}{\underline{0.9873}} \\

+ RAIN-2 & {17.71} & {39.39} & {0.9961} & {14.88} & 34.52 & \textcolor{blue}{\underline{0.9882}} & 13.89 & \textcolor{blue}{\underline{31.19}} & 0.9737 & \textcolor{blue}{\underline{14.39}} & \textcolor{blue}{\underline{27.72}} & {0.9491} & {16.16} & {36.01} & {0.9871} \\

+ RAIN-3 & 17.97 & 39.28 & 0.9960 & 15.00 & {34.54} & {0.9881} & \textcolor{blue}{\underline{13.82}} & \textcolor{blue}{\underline{31.19}} & \textcolor{blue}{\underline{0.9743}} & \textcolor{red}{\textbf{14.21}} & \textcolor{red}{\textbf{27.75}} & \textcolor{blue}{\underline{0.9493}} & 16.30 & 35.95 & {0.9872}\\

+ RAIN-4 & 17.95 & 39.27 & 0.9959 & 14.95 & 34.51 & 0.9878 & \textcolor{red}{\textbf{13.75}} & \textcolor{red}{\textbf{31.23}} & 0.9735 & 14.75 & 27.51 & 0.9469 & 16.31 & 35.96 & 0.9868 \\

+ RAIN-Encoder & 19.29 & 38.81 & 0.9957 & 16.64 & 33.79 & 0.9869 & 15.96 & 30.15 & 0.9719 & 16.40 & 26.72 & 0.9449 & 17.89 & 35.31 & 0.9861 \\

+ RAIN-Decoder & \textcolor{blue}{\underline{17.41}} & \textcolor{blue}{\underline{39.50}} & \textcolor{blue}{\underline{0.9962}} & \textcolor{red}{\textbf{14.32}} & \textcolor{red}{\textbf{34.89}} & \textcolor{red}{\textbf{0.9889}} & 14.18 & 31.01 & \textcolor{red}{\textbf{0.9746}} & {14.75} & 27.60 & \textcolor{red}{\textbf{0.9507}} & \textcolor{blue}{\underline{15.92}} & \textcolor{red}{\textbf{36.12}} & \textcolor{red}{\textbf{0.9877}} \\
\bottomrule

\end{tabular}
\end{center}
\caption{Ablation studies. The numbers in \textcolor{red}{\textbf{red}} and \textcolor{blue}{\underline{blue}} represent the best and second-best performance. }
\label{tab:performance_on_ablation}
\end{table*}

\subsection{Comparison with existing methods}
\label{subsec:comparison}

\noindent
\textbf{Performance on different sub-datasets.} To quantitatively validate our approach, we adopt the evaluation protocols from previous work~\cite{cong2020dovenet,tsai2017deep,cun2020improving}. We first train our model on the whole training set. Then we evaluate the trained model on given testing images by measuring mean square error (MSE) and PSNR score for the synthesized images. The results of all previous methods as well as our RainNet are given in Table~\ref{tab:performance_on_4datasets}. It can be observed that the baseline model attains comparable performance of DIH~\cite{tsai2017deep}. %, and sub-optimal performance among all methods on the Hday2night subset. 
Benefiting from the proposed RAIN module, our RainNet improves the baseline by a reduction of 39.68 in MSE metric, and a performance gain of 2.2 in PSNR for all datasets. Although DoveNet~\cite{cong2020dovenet} is slightly favorable to our approach in Hday2night dataset, our model achieves the best results on HCOCO, HAdobe5k, and HFlickr and outperforms~\cite{cong2020dovenet} by a large margin in average performance.

\noindent
\textbf{Influence of foreground ratios.} We next examine the influence of different foreground ratios on the harmonization models. Following~\cite{cong2020dovenet}, we split the images into three groups according to different foreground ratio ranges, \ie, 0$\%\sim$5$\%$, 5$\%\sim$15$\%$, and 15$\%\sim$100$\%$. We compare the performance by metrics of MSE and fMSE. For fMSE, we only calculate the MSE of the foreground regions. The comparison results are presented in Table~\ref{tab:performance_on_foreground_ratio}. As can be found, on one hand, the model performance in terms of MSE downgrades as the foreground ratios increases while fMSE is less likely to be influenced by foreground ratios. On the other hand, our model outperforms~\cite{cong2020dovenet} by 80.36 in the fMSE metric and improves the performance of the baseline model by 39.68, 339.08 in MSE, fMSE, respectively.

\noindent
\textbf{Qualitative comparisons.} %In Fig.~\ref{fig:comparison1} and Fig.~\ref{fig:comparison2}, 
We proceed to take a closer look at model performance and provide qualitative comparisons with the previous competing methods. From the sample results in Fig.~\ref{fig:comparison1}, it can be easily observed that our method better integrates the foreground objects into the background image, achieving much better visual consistency compared to other methods. For instance, in the second row of Fig.~\ref{fig:comparison1}, the background image is underexposed, while the foreground objects (balloons) are much brighter, leading to unrealistic visual results. Both DIH and DoveNet cannot adjust the foreground to be compatible with the dim backgrounds, while S$^2$AM generates the least realistic result. Our RainNet achieves more photorealistic results with context consistency by adaptively learning the \emph{style} features from the background and applying to the foreground objects. Fig.~\ref{fig:comparison2} gives another three typical samples picked from 99 real composited images evaluated in~\cite{tsai2017deep}. Although there is no ground truth image as a reference, we can still observe significant improvements of visual \emph{style} consistency achieved by our approach.

\begin{figure}
\begin{center}
   % \centering
   \includegraphics[width=0.9\linewidth]{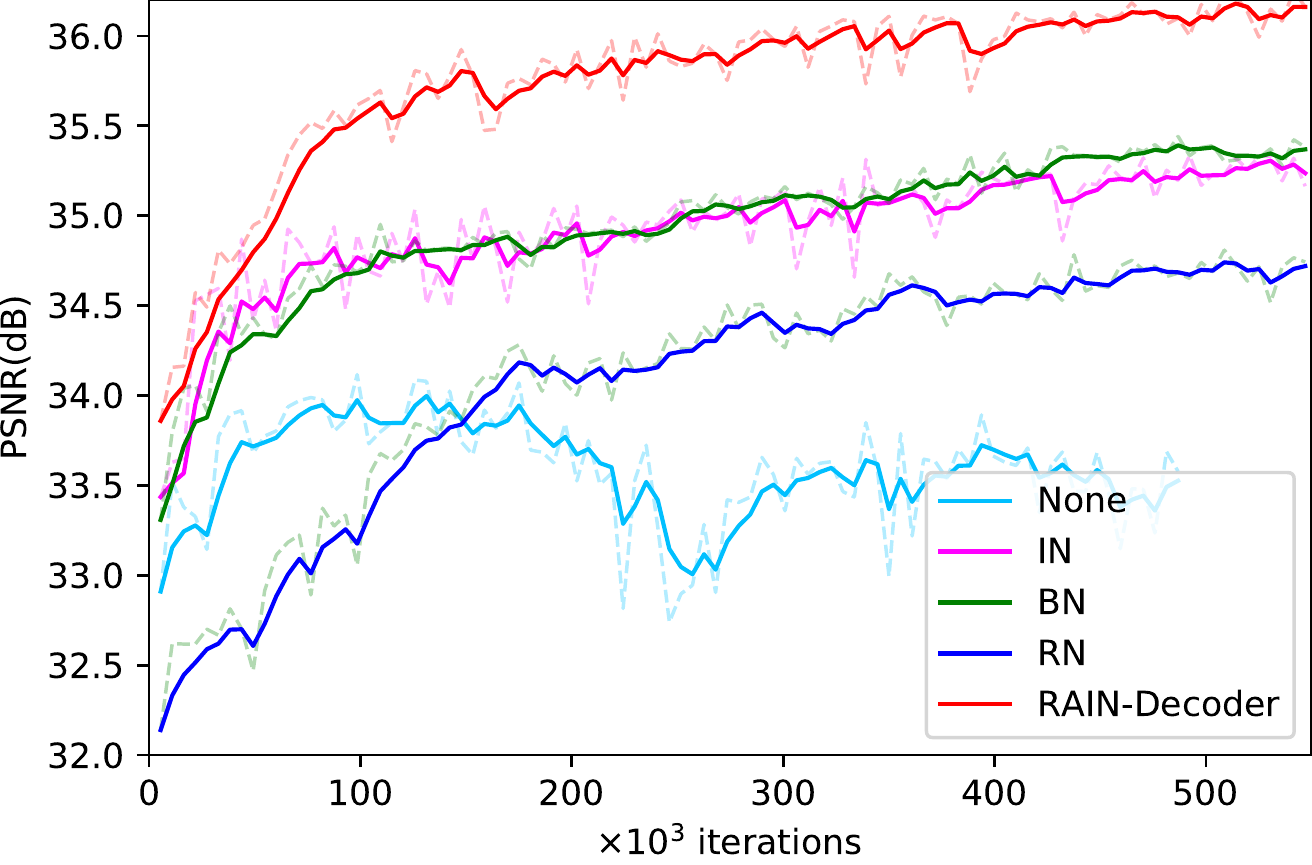}
\end{center}
   \caption{Comparisons of different normalization methods on PSNR metric. Without normalization (labeled by \emph{None}), the model performance heavily deteriorates. }
\label{fig:with_normalization}
\end{figure}
\subsection{Ablation study}
\label{subsec:ablation}
In this section, we conduct comprehensive ablation studies to demonstrate the effectiveness of our RAIN module. Different from Sec.~\ref{subsec:comparison}, we resort to three alternative measures (\ie, foreground L1 norm (fL1), PSNR, and SSIM~\cite{wang2004imagessim}) for quantitative evaluation.

\noindent
\textbf{Efficacy of RAIN.} We first investigate the performance gain brought by our RAIN module compared to other normalization methods, \ie, RN, IN, and BN. To begin with, we apply RN to the baseline model and observe stable model training curves and better performance than that without noralization layers (See in Table~\ref{tab:performance_on_ablation} and Fig.~\ref{fig:with_normalization}). Note that RN only performs batch normalization for the background (foreground) features within all background (foreground) regions, respectively. This operation splits the background and foreground features and prevents the network from propagating information from the background to the foreground, thus cannot generalize well in image harmonization tasks.

We proceed to add IN and BN to the baseline. As can be found in Table~\ref{tab:performance_on_ablation} and Fig.~\ref{fig:with_normalization} (the purple and green curves), the baseline+IN/BN outperforms the baseline method and baseline+RN by a large margin. Potential explanations can be analyzed from two aspects. On one hand, feature normalizing operations can help to stabilize and benefit the training process of deep neural networks, yielding better convergence. On the other hand, performing feature normalization with IN or BN enables the foreground features to be modified by the mean and variance statistically measured from both the foreground features and the background features. Therefore, the model can learn to adjust the visual properties of the foreground objects somehow. 

Furthermore, we replace the normalization layer in the decoder network with RAIN while setting the normalization layer to IN in the encoder, then train the network under the same settings. The results are plotted in Fig.~\ref{fig:with_normalization} (red curve). Obviously, thanks to our novel RAIN module, the model with RAIN-Decoder outperforms other normalization methods and achieves the best performance on average.

\noindent
\textbf{Which layer to add RAIN?} In order to exploit the best implementation strategy for RAIN, we conduct experiments by gradually adding and removing the RAIN layers in the RainNet network. Here we compare several variants that are boosted by RAIN module in different convolutional stages (more variants and comparisons are presented in the supplementary materials). Note that in the middle layers of the generator, the spatial size of convolutional features decreases significantly. For instance, when we resize the foreground mask to 4$\times$4, the valid pixels of the foreground mask are rather rare. Under these circumstances, our RAIN downgrades to Instance Normalization. So we gradually remove RAIN layers from the 4 outermost layers in the encoder and decoder. \textbf{(a) Baseline+RAIN-Decoder}: we add RAIN layer to the decoder and IN to the encoder. \textbf{(b) Baseline+RAIN-Encoder}: in contrast to \textbf{(a)}, we use RAIN a layer only for the encoder and use IN for the decoder. \textbf{(c) Baseline+RAIN-k}: we add k (k=1,2,3,...) RAIN layers to the outermost four layers of the encoder and decoder, and IN to the remaining layers. 

The quantitative comparison results are provided in Table~\ref{tab:performance_on_ablation} and Fig.~\ref{fig:how_normalize}. Our observations can be summarized as follows: 

\begin{figure}
\begin{center}
   % \centering
   \includegraphics[width=0.9\linewidth]{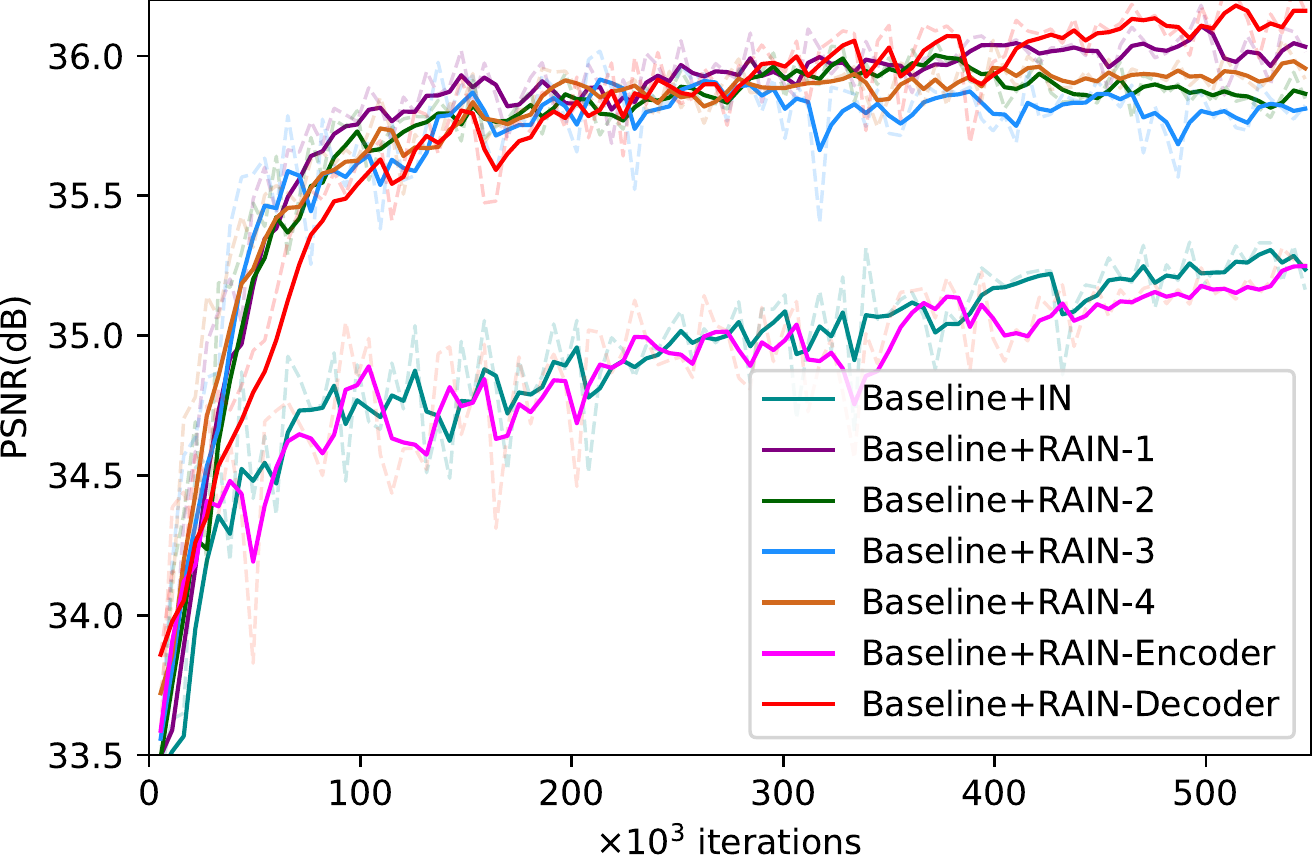}
\end{center}
   \caption{Comparisons of different implementation strategies of RAIN on PSNR metric. }
   \label{fig:how_normalize}
\end{figure}

\noindent
\textbf{1}) Baseline+RAIN-Encoder achieves comparable performance of that with IN, while Baseline+RAIN-Decoder outperforms RAIN-Encoder by a large margin. The differences indicate the better choices of RAIN for the decoder and IN for the encoder. 

\noindent
\textbf{2}) Starting from Baseline+RAIN-Decoder, we decrease the number of RAIN layers in the decoder, while adding as many RAIN layers to the outermost parts of the encoder, \ie, Baseline+RAIN-4. The model attains dropped performance but still better than Baseline+IN. 

\noindent
\textbf{3}) Baseline+RAIN-1 slightly outperforms Baseline+RAIN-2, Baseline+RAIN-3, and Baseline+RAIN-4 by minor improvements. However, when compared to IN, BN, and RN, the improvements brought by our RAIN are significant.

From the experimental results, we conclude that adopting RAIN in the decoder and IN in the encoder or using the similar structure as Baseline+RAIN-$k$ are better choices. One probable reason is that some visual-consistency related features (\eg, color tone, illumination \etc) are likely to be related to the low-level features extracted in the shallow layers of convolutional neural networks, so the layers that are closest to the network’s input and output impose greater impacts on estimation error. Another reason is that the deployment of the RAIN in the symmetrical layers of the encoder and decoder helps the concatenated features have the same mean-variance in the background and foreground regions, which is helpful for the filters to stabilize the training and converge to better performance.

\noindent
\textbf{Adding RAIN to previous work.} To apply RAIN in existing methods, we conduct experiments with DIH~\cite{tsai2017deep}. We first implement DIH (with segmentation branch) in Pytorch~\cite{paszke2017automatic} and then train the basic network. In order to add RAIN to DIH, we replace BN with IN in the encoder, and RAIN with BN in the harmonization decoder. The performance of DIH model reaches to 33.36dB of PSNR while the new model with RAIN achieves 33.84dB (+0.48dB). Detailed illustrations can be found in the supplementary materials.

\begin{table}
\footnotesize
\begin{center}
\begin{tabular}{p{1.3cm}<{\centering}p{0.6cm}<{\centering}p{0.8cm}<{\centering}p{1cm}<{\centering}p{1.1cm}<{\centering}p{1.0cm}<{\centering}}
\toprule
Method & Input & DIH~\cite{tsai2017deep} & S$^2$AM~\cite{cun2020improving} & DoveNet~\cite{cong2020dovenet} & RainNet\\
\midrule
Total votes & 113 & 203 & 193 & 226 & \textbf{354} \\
Preference & 10.4$\%$ & 18.6$\%$ & 17.7$\%$ & 20.8$\%$ & \textbf{32.5}$\%$ \\
\bottomrule
\end{tabular}
\end{center}
\caption{Comparisons between our method and other competing methods under user study.}
\label{tab:user_study}
\end{table}

% \noindent
\subsection{User study} 
\label{subsec:user_study}
Table~\ref{tab:user_study} shows the user evaluation results on real-world composited images collected by DIH~\cite{tsai2017deep}. Specifically, we invited 11 volunteers to rate and choose the most realistic harmonized images from 5 given images. Those 5 images include the original composite image and its corresponding 4 harmonized versions created by DIH, S$^2$AM, DoveNet, and Ours. We randomly shuffle the displaying order of 5 images to ensure that the users do not know which model each image belongs to. Each user is asked to evaluate for the whole set (99 images). As shown in the Table~\ref{tab:user_study}, RainNet attains more votes than the rest, which demonstrates the effectiveness of the proposed approach.

\subsection{Discussions and limitations}
\label{subsec:limitation_discussion}
\noindent\textbf{Discussions.} Obviously, benefiting from RAIN module, RainNet achieves a higher PSNR score and lower estimation error than previous DoveNet~\cite{cong2020dovenet} by 1.37dB and 12.07, respectively. Although we found that parts of these improvements are attributed to our generator settings, in which we only learn to modify the foreground image and copy the background pixels from the input, thus reducing the error of the background, we attain lower foregroud estimation errors (fMSE). fMSE is fair for all methods. Furthermore, comparing to IN, RainNet remarkably improves the performance of a baseline model and achieves the best scores on average, which demonstrates the superiority of the proposed RAIN module. 

% Despite the success of our method in reducing visual \emph{sytle} discrepancy between the foreground image and the background images, 
\noindent\textbf{Limitations.} Despite the improvements, our proposed approach still faces with two major confusions. First, it is not very clear why applying RAIN only in the encoder brings little improvement. Second, our model will soften the sharp foreground object and reduce the visual style discrepancy in the samples with dark background and sharp foreground objects. Future investigation in these issues should be required. 

\section{Conclusion}
\label{sec:conclusion}
In this paper, we propose to solve the visual \emph{style} inconsistency problem in image harmonization and present a simple yet effective Region-aware Adaptive Instance Normalization (RAIN) module, which outperforms previous normalization methods by a large margin. We have also exploited the best implementation choice of RAIN for the baseline network. Moreover, we demonstrate the efficacy of RAIN by applying RAIN into existing networks, \eg, DIH, and observe performance gains over these models.

%Both quantitative and qualitative experiments were conducted to verify the efficacy of the proposed approach. 

% \noindent
% \textbf{Acknowledgements.}
% This work was supported in part by MoE-China Mobile Research Fund Project under Grant MCM20180702, National Key R\&D Project of China under Grant 2019YFB1802701, and Shanghai Key Laboratory of Digital Media Processing and Transmissions. 

% \clearpage

{\small
\bibliographystyle{ieee_fullname}
\bibliography{egbib}
}

% \newpage

\appendix

{\Large
\noindent
\newline
\textbf{Supplementary Material}}
\section{Overview}
\label{sec:overview}

In this supplementary, we provide implementation details in Sec.~\ref{sec:implementation}, including the detailed architecture of attention block, and model training objective. We also conduct more ablation studies to exploit better applying strategy of the proposed RAIN method (Sec.~\ref{sec:more_ablation}). More comparison results on real composite images are presented in Sec.~\ref{sec:real_composite_results}. Finally, we discuss the failure case in Sec.~\ref{sec:failure_cases}

\section{Implementation Details}
\label{sec:implementation}
% \begin{figure*}[t]
% \begin{center}
% % \fbox{\rule{0pt}{2in} \rule{0.9\linewidth}{0pt}}
%    % \centering
%    \includegraphics[width=0.9\linewidth]{graph/model-Generator.pdf}
% \end{center}
%    \caption{Overview of the proposed generator. We provide a detailed structure of our RainNet to ensure better understanding and repdeterioratingroducibility. The bottom legend: Conv.= Convolution, Trans. = Transposed. }
% \label{fig:model_attentioned}
% \end{figure*}

%------------------------------------------------------------------------
\subsection{Attention Block}
\label{subsec:attention_block}
\begin{figure}[!htbp]
\begin{center}
   % \centering
   \includegraphics[width=1\linewidth]{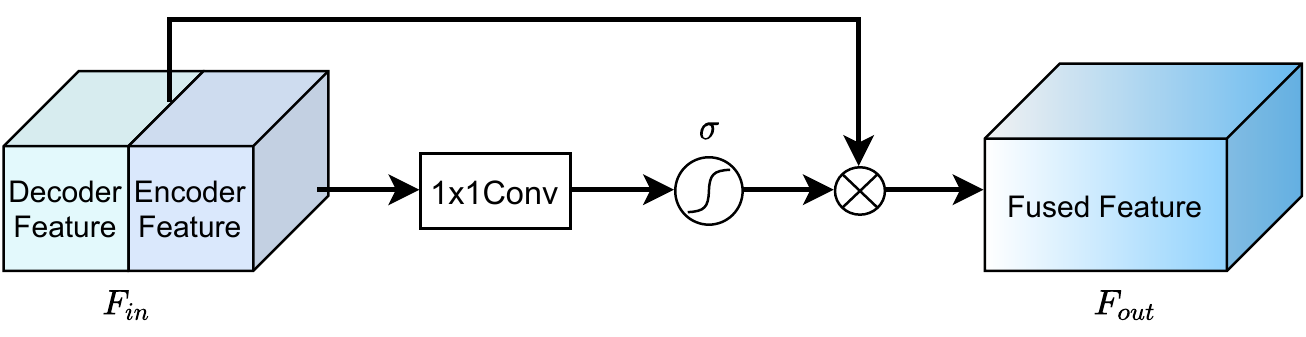}\\
\end{center}
   \caption{Illustration of the adopted attention block. }
\label{fig:attention_block}
\end{figure}

Attention block has been proven to  bring noticeable improvements to the simple U-Net architecutre~\cite{cong2020dovenet,cun2020improving}. Following the prior work, we add three attention blocks in the decoder part for baseline network (the structure of generator is presented in Section 3 of the main paper). The detailed structure of attention block is presented in Fig.~\ref{fig:attention_block}. 

Specifically, in each attention block, we take the concatenation of the encoder feature and the decoder feature $F_{in}\in{\mathbb{R}^{C\times H\times W}}$ as the input of the block. To fuse the concatenated features, we use an 1$\times$1 convolutional layer and a Sigmoid activation function $\sigma$ to acquire coefficients map, which is denoted as $W\in{\mathbb{R}^{C\times H\times W}}$. Then we acquired the modulated feature $F_{out}$ by multiplying the concatenated features by the map in element-wise manner:
\begin{equation}
\label{equ:exciting}
F_{out} = W\circ F_{in},
\end{equation}
where $\circ$ denotes the element-wise multiplication.

%-------------------------------------------------------------------------
\subsection{Improving Image Composites}
\label{subsec:improving_composite}
% Consider a foreground image and a background image as $I_{f}$ and $I_{b}$ respectively. The foreground mask is denoted by $M$, which indicates the region to be harmonized in composite image $I_{c}$. Accordingly, the background mask is $\bar{M}=1-M$. The object composition process is formulated as $I_{c} = M\circ I_{f} + (1-M) \circ I_{b}$, 
% \begin{equation}
% \label{equ:image_composition}
% I_{c} = M\circ I_{f} + (1-M) \circ I_{b}, 
% \end{equation}
% where $\circ$ is Hadamard product. 

\begin{table*}[!tp]
\small
\begin{center}
\begin{tabular}{|c|c|c|c|c|c|c|c|c|c|c|c|c|c|c|c|}
\hline
\multirow{2}{*}{Type} & \multirow{2}{*}{Method} & \multicolumn{14}{c|}{Index of feature normalization layer} \\
\cline{3-16}
& & 1 & 2 & 3 & 4 & 5 & 6 & 7 & 8 & 9 & 10 & 11 & 12 & 13 & 14 \\
\hline
% \hline\hline
% \multirow{8}{*}{I} & Baseline & - & - & - & - & - & - & - & - & - & - & - & - & - & -  \\
% & Baseline + IN & IN & IN & IN & IN & IN & IN & IN & IN & IN & IN & IN & IN & IN & IN\\
\multirow{6}{*}{I}& Baseline + RAIN-Decoder-1 & IN & IN & IN & IN & IN & IN & IN & IN & IN & IN & IN & IN & IN & \textcolor{red}{\textbf{R}} \\
& Baseline + RAIN-Decoder-2 & IN & IN & IN & IN & IN & IN & IN & IN & IN & IN & IN & IN & \textcolor{red}{\textbf{R}} & \textcolor{red}{\textbf{R}} \\
& Baseline + RAIN-Decoder-3 & IN & IN & IN & IN & IN & IN & IN & IN & IN & IN & IN & \textcolor{red}{\textbf{R}} & \textcolor{red}{\textbf{R}} & \textcolor{red}{\textbf{R}} \\
& Baseline + RAIN-Decoder-4 & IN & IN & IN & IN & IN & IN & IN & IN & IN & IN & \textcolor{red}{\textbf{R}} & \textcolor{red}{\textbf{R}} & \textcolor{red}{\textbf{R}} & \textcolor{red}{\textbf{R}} \\
& Baseline + RAIN-Decoder & IN & IN & IN & IN & IN & IN & IN & \textcolor{red}{\textbf{R}} & \textcolor{red}{\textbf{R}} & \textcolor{red}{\textbf{R}} & \textcolor{red}{\textbf{R}} & \textcolor{red}{\textbf{R}} & \textcolor{red}{\textbf{R}} & \textcolor{red}{\textbf{R}} \\
& Baseline + RAIN-Encoder & \textcolor{red}{\textbf{R}} & \textcolor{red}{\textbf{R}} & \textcolor{red}{\textbf{R}} & \textcolor{red}{\textbf{R}} & \textcolor{red}{\textbf{R}} & \textcolor{red}{\textbf{R}} & \textcolor{red}{\textbf{R}} & IN & IN & IN & IN & IN & IN & IN\\
\hline\hline
\multirow{6}{*}{II}& Baseline + RAIN-1 & \textcolor{red}{\textbf{R}} & IN & IN & IN & IN & IN & IN & IN & IN & IN & IN & IN & IN & \textcolor{red}{\textbf{R}} \\
& Baseline + RAIN-2 & \textcolor{red}{\textbf{R}} & \textcolor{red}{\textbf{R}} & IN & IN & IN & IN & IN & IN & IN & IN & IN & IN & \textcolor{red}{\textbf{R}} & \textcolor{red}{\textbf{R}} \\
& Baseline + RAIN-3 & \textcolor{red}{\textbf{R}} & \textcolor{red}{\textbf{R}} & \textcolor{red}{\textbf{R}} & IN & IN & IN & IN & IN & IN & IN & IN & \textcolor{red}{\textbf{R}} & \textcolor{red}{\textbf{R}} & \textcolor{red}{\textbf{R}} \\
& Baseline + RAIN-4 & \textcolor{red}{\textbf{R}} & \textcolor{red}{\textbf{R}} & \textcolor{red}{\textbf{R}} & \textcolor{red}{\textbf{R}} & IN & IN & IN & IN & IN & IN & \textcolor{red}{\textbf{R}} & \textcolor{red}{\textbf{R}} & \textcolor{red}{\textbf{R}} & \textcolor{red}{\textbf{R}} \\
& Baseline + RAIN-5 & \textcolor{red}{\textbf{R}} & \textcolor{red}{\textbf{R}} & \textcolor{red}{\textbf{R}} & \textcolor{red}{\textbf{R}} & \textcolor{red}{\textbf{R}} & IN & IN & IN & IN & \textcolor{red}{\textbf{R}} & \textcolor{red}{\textbf{R}} & \textcolor{red}{\textbf{R}} & \textcolor{red}{\textbf{R}} & \textcolor{red}{\textbf{R}} \\
& Baseline + RAIN-6 & \textcolor{red}{\textbf{R}} & \textcolor{red}{\textbf{R}} & \textcolor{red}{\textbf{R}} & \textcolor{red}{\textbf{R}} & \textcolor{red}{\textbf{R}} & \textcolor{red}{\textbf{R}} & IN & IN & \textcolor{red}{\textbf{R}} & \textcolor{red}{\textbf{R}} & \textcolor{red}{\textbf{R}} & \textcolor{red}{\textbf{R}} & \textcolor{red}{\textbf{R}} & \textcolor{red}{\textbf{R}} \\
\hline\hline
\multirow{3}{*}{III}& Baseline + RAIN-Inner-3 & IN & IN & IN & IN & \textcolor{red}{\textbf{R}} & \textcolor{red}{\textbf{R}} & \textcolor{red}{\textbf{R}} & \textcolor{red}{\textbf{R}} & \textcolor{red}{\textbf{R}} & \textcolor{red}{\textbf{R}} & IN & IN & IN & IN\\
& Baseline + RAIN-Inner-4 & IN & IN & IN & \textcolor{red}{\textbf{R}} & \textcolor{red}{\textbf{R}} & \textcolor{red}{\textbf{R}} & \textcolor{red}{\textbf{R}} & \textcolor{red}{\textbf{R}} & \textcolor{red}{\textbf{R}} & \textcolor{red}{\textbf{R}} & \textcolor{red}{\textbf{R}} & IN & IN & IN\\
& Baseline + RAIN-Inner-5 & IN & IN & \textcolor{red}{\textbf{R}} & \textcolor{red}{\textbf{R}} & \textcolor{red}{\textbf{R}} & \textcolor{red}{\textbf{R}} & \textcolor{red}{\textbf{R}} & \textcolor{red}{\textbf{R}} & \textcolor{red}{\textbf{R}} & \textcolor{red}{\textbf{R}} & \textcolor{red}{\textbf{R}} & \textcolor{red}{\textbf{R}} & IN & IN\\
\hline
\end{tabular}
\end{center}
\caption{Designing choices of RAIN. IN: Instance Normalization, \textcolor{red}{\textbf{R}}: RAIN. }
\label{tab:applying_choice_of_RAIN}
\end{table*}

In this paper, we define the composite image as $I_{c}$, the foreground mask as $M$. The harmonization model is denoted by $G$, and the harmonized image by $\hat{I} = G(I_{c}, M)$. Our aim is to optimize the model $G$ to make $\hat{I}$ close to the ground truth image $I$ by a reconstruction loss: 
\begin{equation}
\label{equ:rec}
\mathcal{L}_{rec}(G, I, I_{c}, M) = \|G(I_{c}, M) - I \|_{1}.
\end{equation}

Due to the widespread applications of adversarial training in many computer vision tasks, we also adopt adversarial training method and follow the training strategy in~\cite{cong2020dovenet,cun2020improving}. The adversarial loss can be written as follows:
\begin{equation}
\label{equ:adv_loss_d_hinge}
\begin{split}
\mathcal{L}_{adv}(D, I, \hat{I}) &= \mathbb{E}_{I}[\max(0, 1 - D(I))] \\
&+\mathbb{E}_{\hat{I}}[\max(0, 1+D(\hat{I}))], \\
\end{split}
\end{equation}
and
\begin{equation}
\label{equ:adv_loss_g_hinge}
\begin{split}
\mathcal{L}_{adv}(G, I_{c}, M) =& -\mathbb{E}_{I_{c}}[D(G(I_{c}, M))], 
\end{split}
\end{equation}
where $D$ tries to distinguish between natural-realistic images $I$ and harmonized samples $\hat{I}$, while $G$ aims to generate samples that look similar to the real observations. Introducing adversarial loss can, in theory, learn the model $G$ that generate images as realistic as the real~\cite{goodfellow2014generative,isola2017image}. 

Besides the global discriminator, we also adopt the setting of domain verification loss~\cite{cong2020dovenet}, which has been proved to bring modest improvements for image harmonization. Specifically, we construct real and fake samples by grouping image pairs of $(I\circ M, I\circ (1-M))$ and $(\hat{I}\circ M, \hat{I}\circ (1-M))$, respectively. To perform domain-oriented optimization, we first utilize a domain encoder $E_{D}$ to obtain the feature representations of the foreground image and the background image. We denote the feature representations as $l_{f}$ and $l_{b}$, respectively. % (where $l_{f}, l_{b}=E_{D}(I, M)$). 
Equally, $\hat{l}_{f}$ and $\hat{l}_{b}$ are extracted from harmonized image $\hat{I}$ by the same domain encoder. To acquire domain verification loss, following~\cite{cong2020dovenet}, we use one more domain discriminator $D_{v}$ which incorporate the domain encoder $E_{D}$ and measure the similarity of $l_{f}$ and $l_{b}$ by:
\begin{equation}
D_{v}(I, M)=l_{f}\cdot l_{b},
\end{equation}
where $\cdot$ means the inner product of two vectors.

Afterward, we measure the domain verification loss as follows: 
\begin{equation}
\label{equ:ver_loss_d_hinge}
\begin{split}
\mathcal{L}_{v}(D_{v}, I, \hat{I}, M) &= \mathbb{E}_{I}[\max(0, 1-D_{v}(I, M))] \\
&+ \mathbb{E}_{\hat{I}}[\max(0, 1+D_{v}(\hat{I}, M))],
\end{split}
\end{equation}
\begin{equation}
\label{equ:ver_loss_g_hinge}
\begin{split}
\mathcal{L}_{v}(G, I_{c}, M) &= -\mathbb{E}_{I_{c}}[D_{v}(G(I_{c}, M), M)].
\end{split}
\end{equation}
By using domain verification loss, the discriminator is encouraged to distinguish similar domain features for positive foreground-background pairs from negative foreground-background pairs. 

In our experiments, $D$ and $D_{v}$ share the same structure as~\cite{cong2020dovenet}, and we apply the well-know spectral normalization~\cite{miyato2018spectral} for two discriminators to stabilize training procedure. The domain encoder utilizes Partial Convolutions~\cite{liu2018image} to extract domain code for regions with irregular shape, avoiding information leakage from unmasked regions.

Our full objective is: 
\begin{equation}
\label{equ:full_objective_d}
% \begin{split}
\mathcal{L}(D, D_{v}, I, \hat{I}, M) = \lambda_{1}\mathcal{L}_{adv}(D, I, \hat{I})
+ \lambda_{2}\mathcal{L}_{v}(D_{v}, I, \hat{I}, M),
% \end{split}
\end{equation}
\begin{equation}
\label{equ:full_objective_g}
\begin{split}
\mathcal{L}(G, I, I_{c}, M) &= \lambda_{1}\mathcal{L}_{adv}(G, I_{c}, M) + \lambda_{2}\mathcal{L}_{v}(G, I_{c}, M) \\
&+ \lambda_{3}\mathcal{L}_{rec}(G, I, I_{c}, M), 
\end{split}
\end{equation}
where $\lambda_{1}=\lambda_{2}=1$, and $\lambda_{3}=100$. 
% Note that our RAIN layer only performs feature normalization after convolutional layers. Therefore, we keep the weights of each loss as same as [10]. 
% In our experiment, we solve this problem by learning a min-max optimizing task:
% \begin{equation}
% \label{equ:min_max}
% G^{*} = \arg \min_{G}\max_{D, D_{v}} \mathcal{L}(D, D_{v}, G)
% \end{equation}

\begin{figure*}
\begin{center}
\subfigure[]{
	\includegraphics[width=0.45\linewidth]{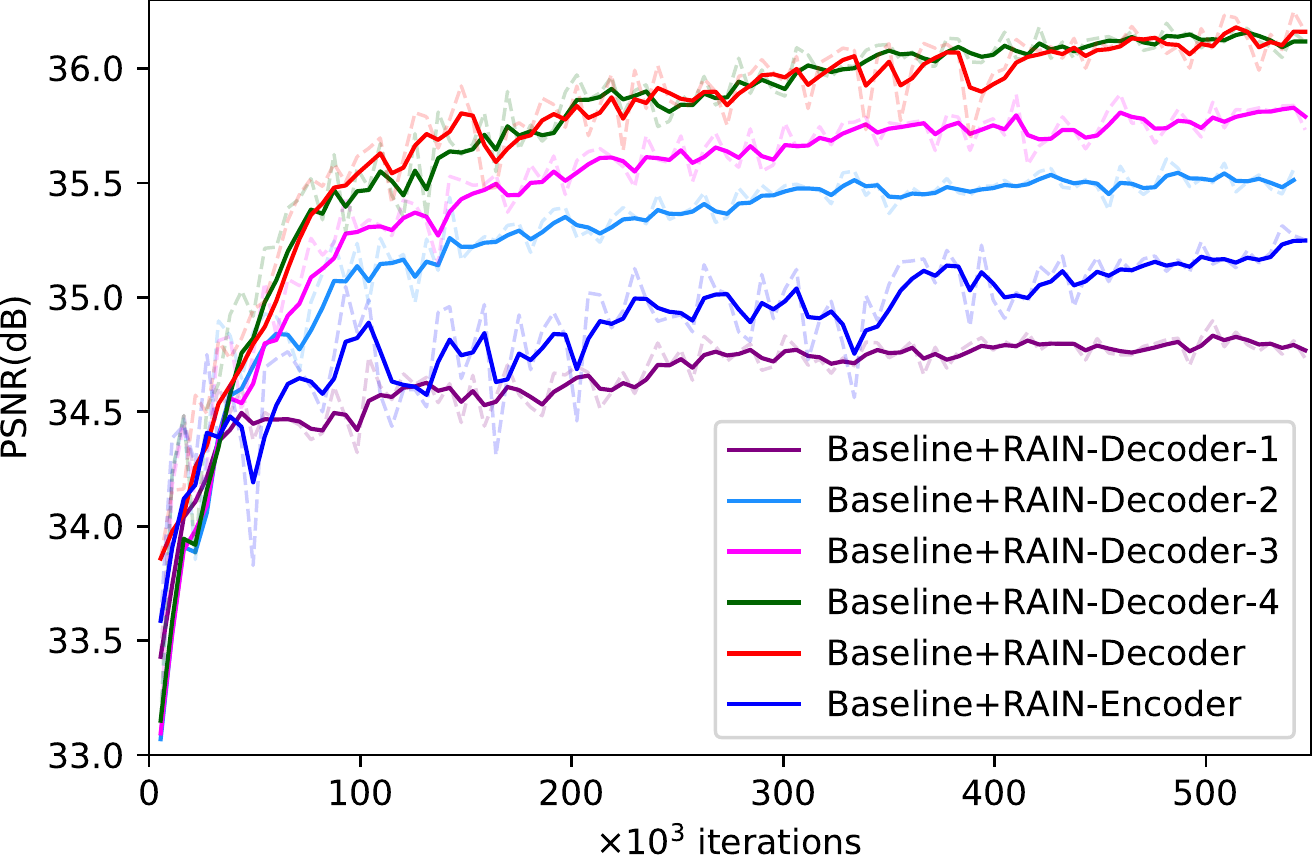}
	\label{fig:decoder_layers}}\hspace{0.4cm}
\subfigure[]{
   	  \includegraphics[width=0.45\linewidth]{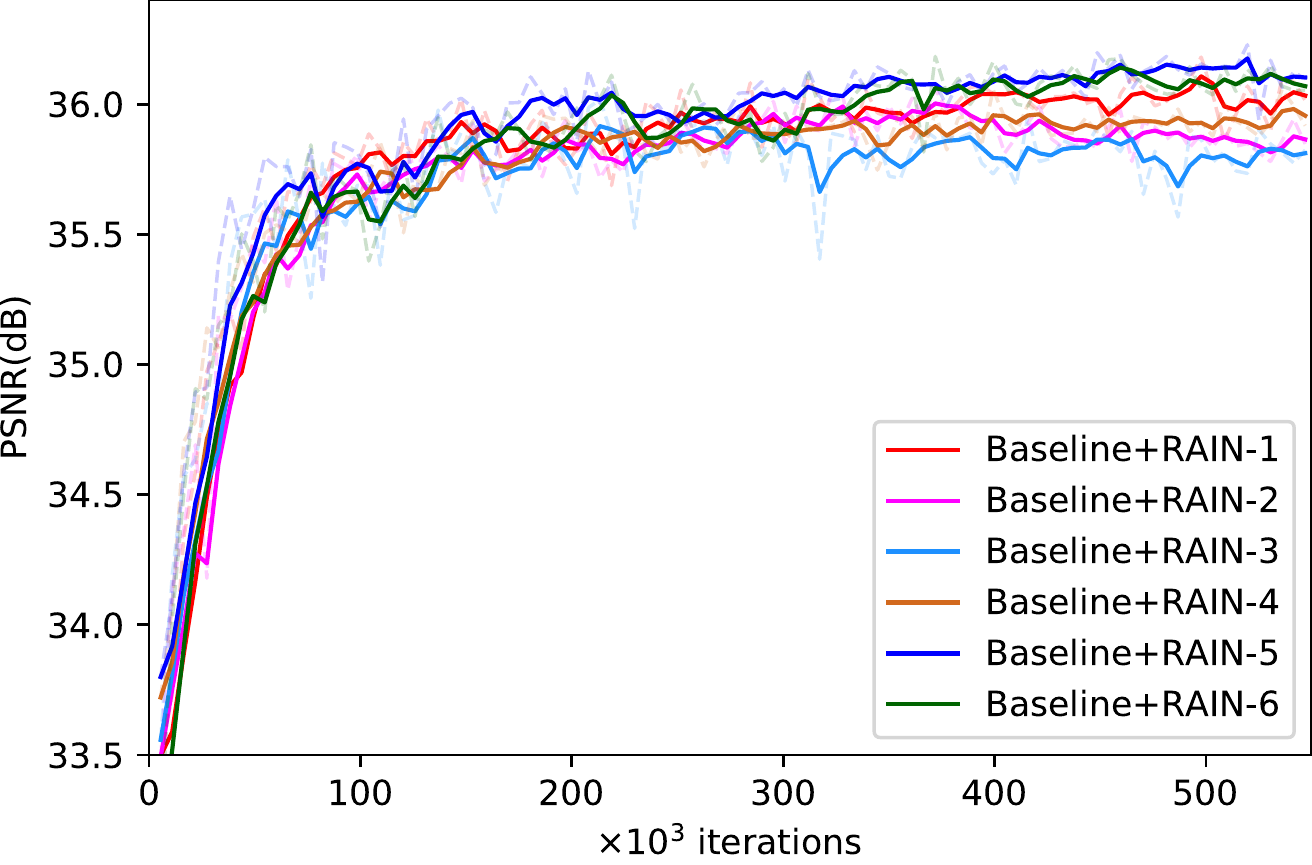}
      \label{subfig:symmetric_structure}}\vspace{0cm}\\\hspace{0.2cm}
\subfigure[]{
	   \includegraphics[width=0.45\linewidth]{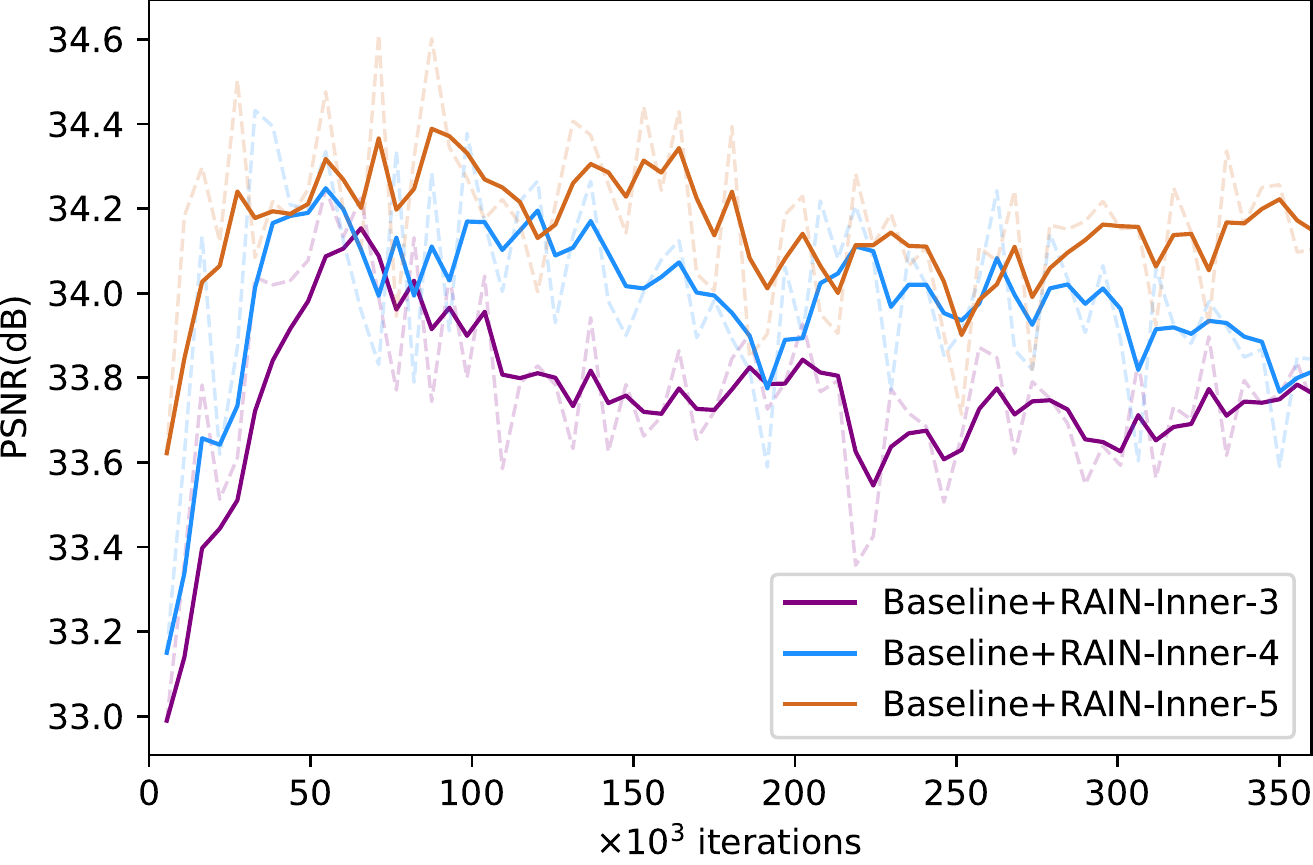}
	   \label{subfig:middle_layers}}\hspace{0.25cm}
\subfigure[]{
	   \includegraphics[width=0.465\linewidth]{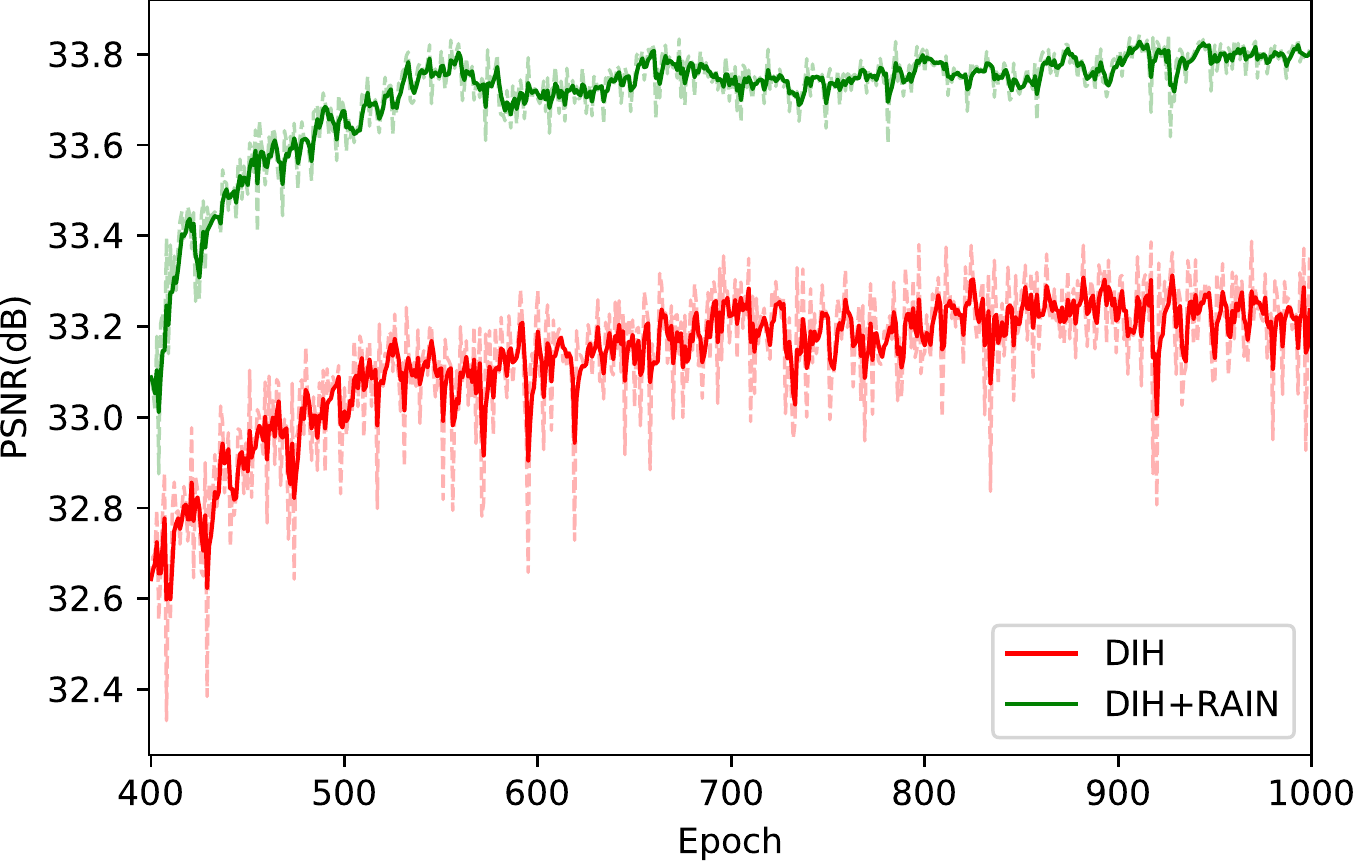}
	   \label{fig:dih_rain}}
\end{center}
\caption{
  Convergence curves on PSNR metric. (a) Type I. We only add RAIN layers to the decoder or encoder. (b) Type II: we add RAIN modules to the outermost layers of the network; (c) Type III: we add RAIN modules to the innermost layers of the network. (d) Convergence curves of DIH~\cite{tsai2017deep} and DIH+RAIN on PSNR metric. Better viewed in color with zoom in.}
\label{fig:ablation_curves}
\end{figure*}

\section{More ablation studies}
\label{sec:more_ablation}
In this section, we conduct more experiments to validate the efficacy of our method. Theoretically, our RAIN module can be applied in any layers of the basic network. In this section, we train our baseline with different designing strategies of applying our RAIN module. 

As presented in Table~\ref{tab:applying_choice_of_RAIN}, we exploit better implementations of RAIN module by designing three main types of structures of the basic network: 
\textbf{I}) we gradually replace IN with RAIN in the decoder or encoder; \textbf{II}) we add RAIN modules to the outermost layers of the network; \textbf{III}) we add RAIN modules to the innermost layers of the network. We conduct these experiments with fixed random seed for better reproduction. The convergence results are presented in Fig.~\ref{fig:ablation_curves}. 

From Fig.~\ref{fig:decoder_layers}, it is obvious that more RAIN layers in the decoder brings more stable training process and better convergence performance. When we only add one RAIN layer at the last normalization layer of the network, \ie, Baseline+RAIN-Decoder-1, we attain the least PSNR results (purple curve). As we add more RAIN layers to the decoder, we obtain noticeable improvements. Another interesting conclusion is that adding more RAIN layers to the decoder brings no benifits when we have already added four RAIN layers in the decoder(green curve and red curve). This may be ascribed to the reasons that when the feature size is small enough, \eg, 4$\times$4 or 8$\times$8, our RAIN module will equal to IN. Therefore, equal performances of Baseline+RAIN-Decoder and Baseline+RAIN-Decoder-4 are observed in our experiments.  

In Fig.~\ref{subfig:symmetric_structure} and Fig.~\ref{subfig:middle_layers}, we visualize the convergence curves of methods within type II and III. %We have the following observations. 
It is clear that using symmetric normalization method for the network benefits the model optimizing process and leads to better convergent performance. Specifically, from subfigure~\ref{subfig:symmetric_structure}, Baseline+RAIN-5 and Baseline+RAIN-6 outperform other methods, while Baseline+RAIN-1 performs slightly better than the Baseline+RAIN-2/3/4. %We subscribe the performance gain to the symmetric setting. 

In Fig.~\ref{subfig:middle_layers}, we visualize the convergent curves of those methods with RAIN modules inserted in the middle part of the network. It can be observed that Baseline+RAIN-Inner-5 is much better than Baseline+RAIN-Inner-3 but much worse than Baseline+RAIN-5. To analyze the observation, note that the visual style defined in this work is close to image visual properties, including illumination, color temperature, saturation, hue, and texture, \etc. In other words, visual properties in image harmonization task are more related to low-level feature  representations learnd by convolutional network in the first few layers of the encoder and the last few layers in the decoder. Therefore, adding the same amount of RAIN layers to the middle layers of the baseline network is less competitive than adding to the outermost layers. 

% \begin{figure}[!htbp]
% \begin{center}
%    % \centering
%    \includegraphics[width=1\linewidth]{supplemental/DIH_with_RAIN.pdf}
% \end{center}
%    \caption{Convergence curves of DIH~\cite{tsai2017deep} and DIH+RAIN on PSNR metric. }
% \label{fig:dih_rain}
% \end{figure}

\noindent
\textbf{Adding RAIN to previous work.}
We first re-implemented DIH in PyTorch and pretrain the whole model for the first 400 epochs. Then we freeze the segmentation branch and optimize the encoder and harmonization branch for another 600 epochs. To add RAIN to DIH, we replace BN with IN in the encoder, and BN with RAIN in the harmonization decoder. Note that we only predict the foreground objects like RainNet does. In Fig.~\ref{fig:dih_rain}, we present the performance curve of DIH and its variant. It can be easily conclude that RAIN module stabilizes the optimizing process and brings significant improvements to existing network.

\section{Results on real composite images}
\label{sec:real_composite_results}
\begin{figure*}
\begin{center}
   % \centering
   \includegraphics[width=1\linewidth]{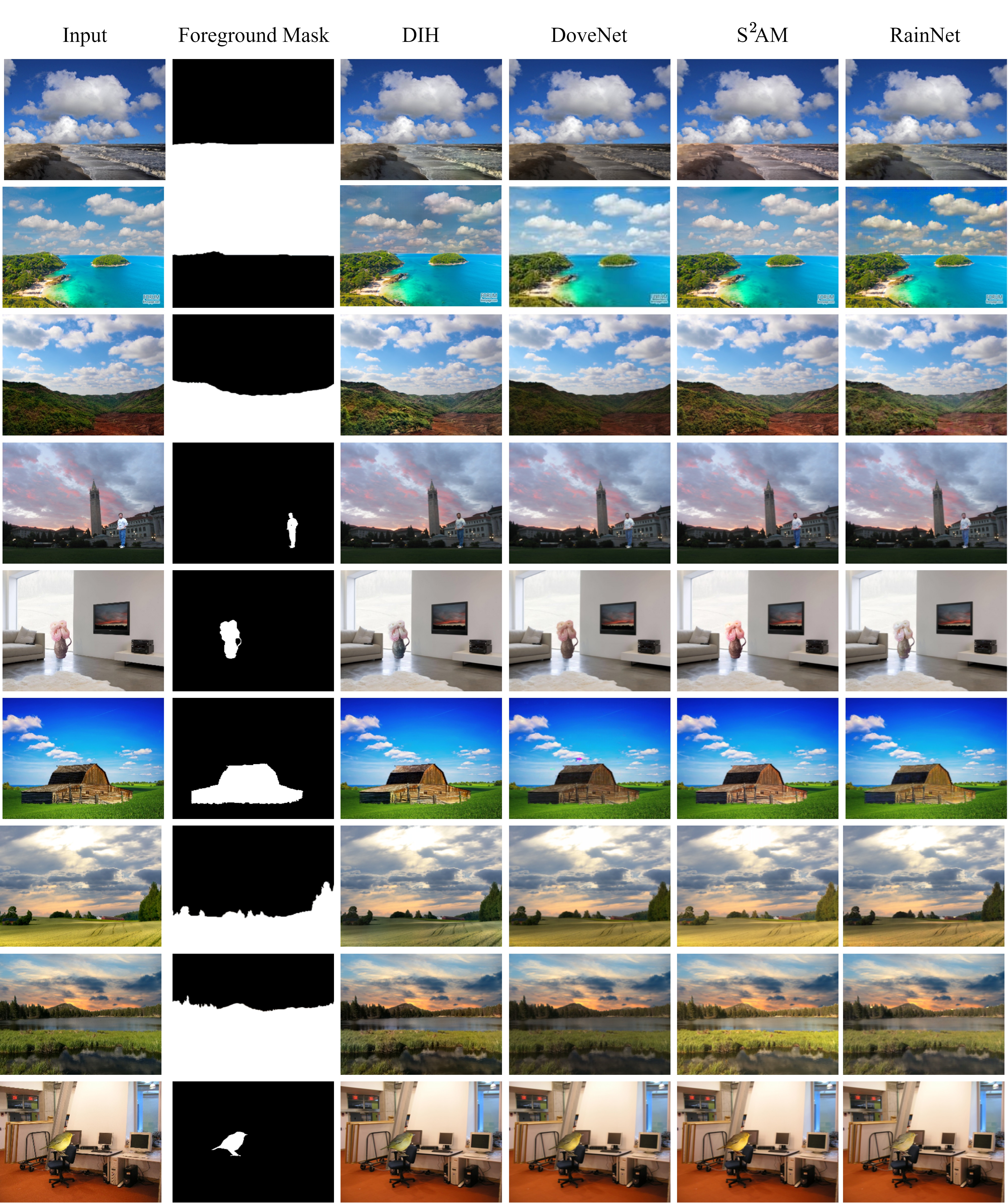}
\end{center}
   \caption{\textbf{Example results on real composite images.}. We present real composite images, foreground mask, three state-of-the-art methods, and the proposed model. The samples are taken from the testing dataset of~\cite{tsai2017deep}. Our method achieves better harmonized visual results than competing methods. }
   \label{fig:result1}
\end{figure*}

\begin{figure*}
\begin{center}
   % \centering
   \includegraphics[width=1\linewidth]{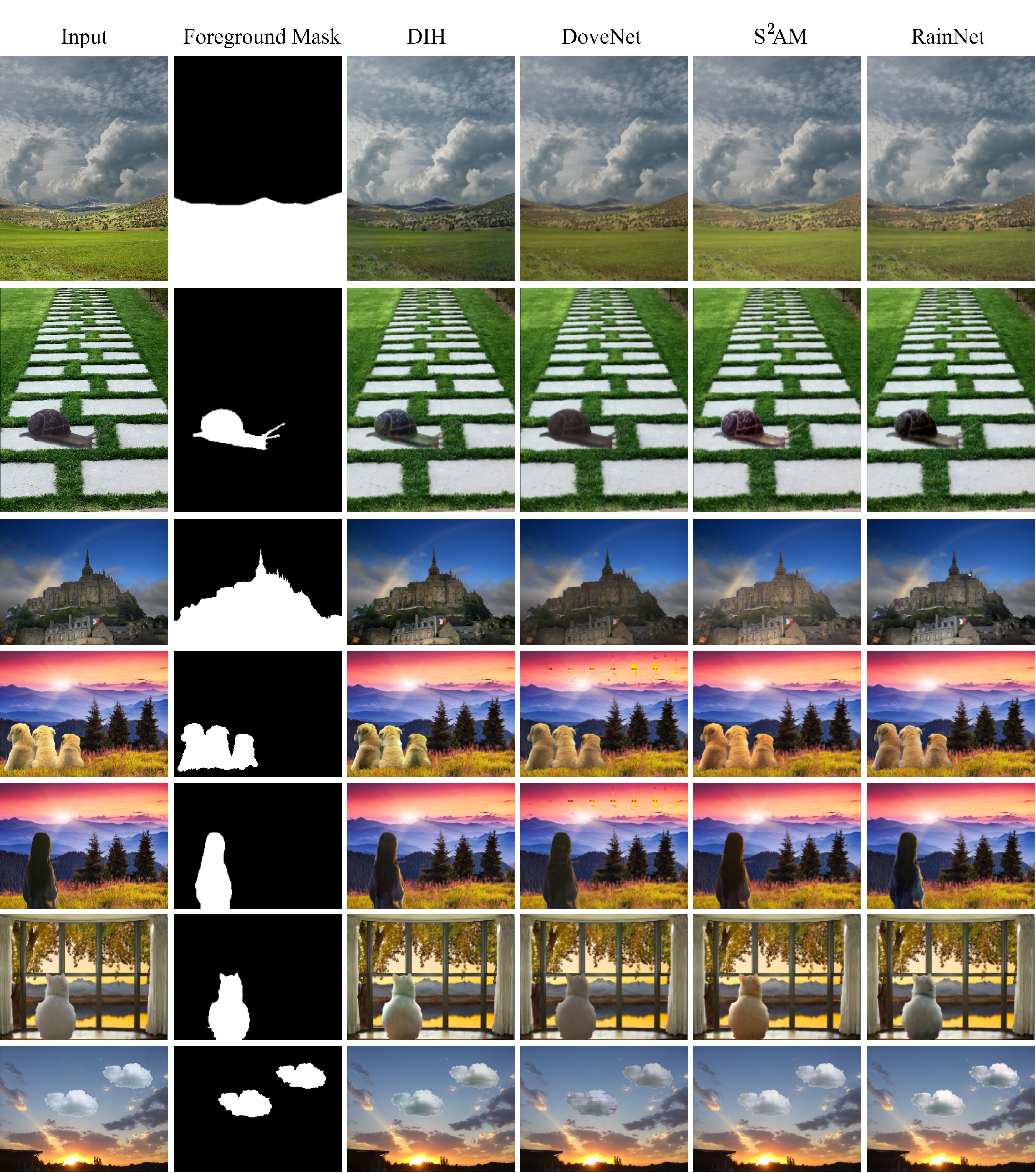}
\end{center}
   \caption{\textbf{Example results on real composite images.}. We present real composite images, foreground mask, three state-of-the-art methods, and the proposed model. The samples are taken from the testing dataset of~\cite{tsai2017deep}. Our method achieves better harmonized visual results than competing methods. }
   \label{fig:result2}
\end{figure*}

\begin{figure*}
\begin{center}
   % \centering
   \includegraphics[width=1\linewidth]{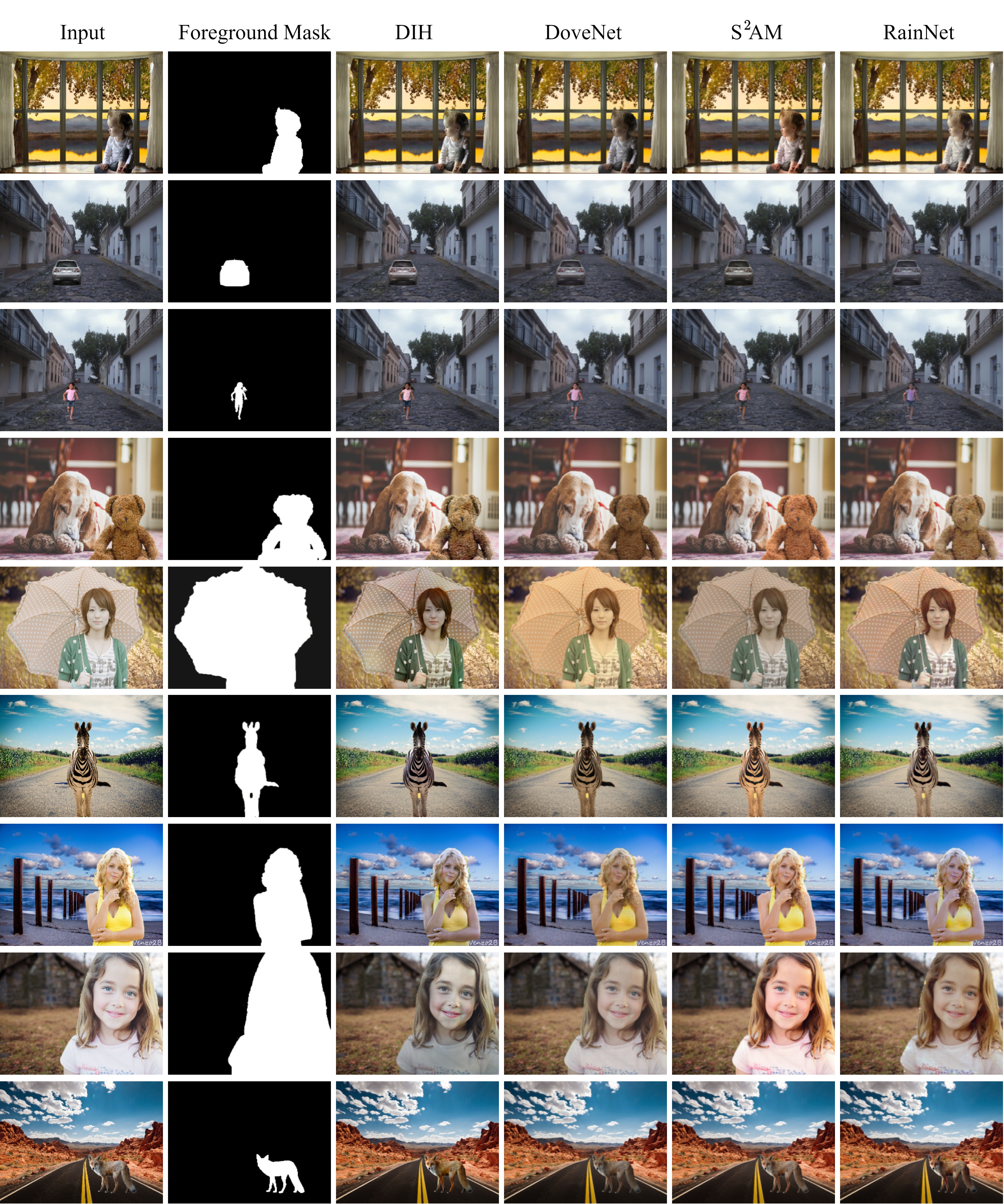}
\end{center}
   \caption{\textbf{Example results on real composite images.}. We present real composite images, foreground mask, three state-of-the-art methods, and the proposed model. The samples are taken from the testing dataset of~\cite{tsai2017deep}. Our method achieves better harmonized visual results than competing methods. }
   \label{fig:result3}
\end{figure*}

In this section, we present the sample results of real composite image used in~\cite{tsai2017deep} and \cite{cong2020dovenet} and compare our method to other competing methods in Fig.~\ref{fig:result1}, \ref{fig:result2} and \ref{fig:result3}. As can be found, our method chieves better visual consistency between the foreground and the background images and outperforms other methods in most cases.

\section{Failure case}
\label{sec:failure_cases}
\begin{figure}
\begin{center}
   % \centering
   \includegraphics[width=1\linewidth]{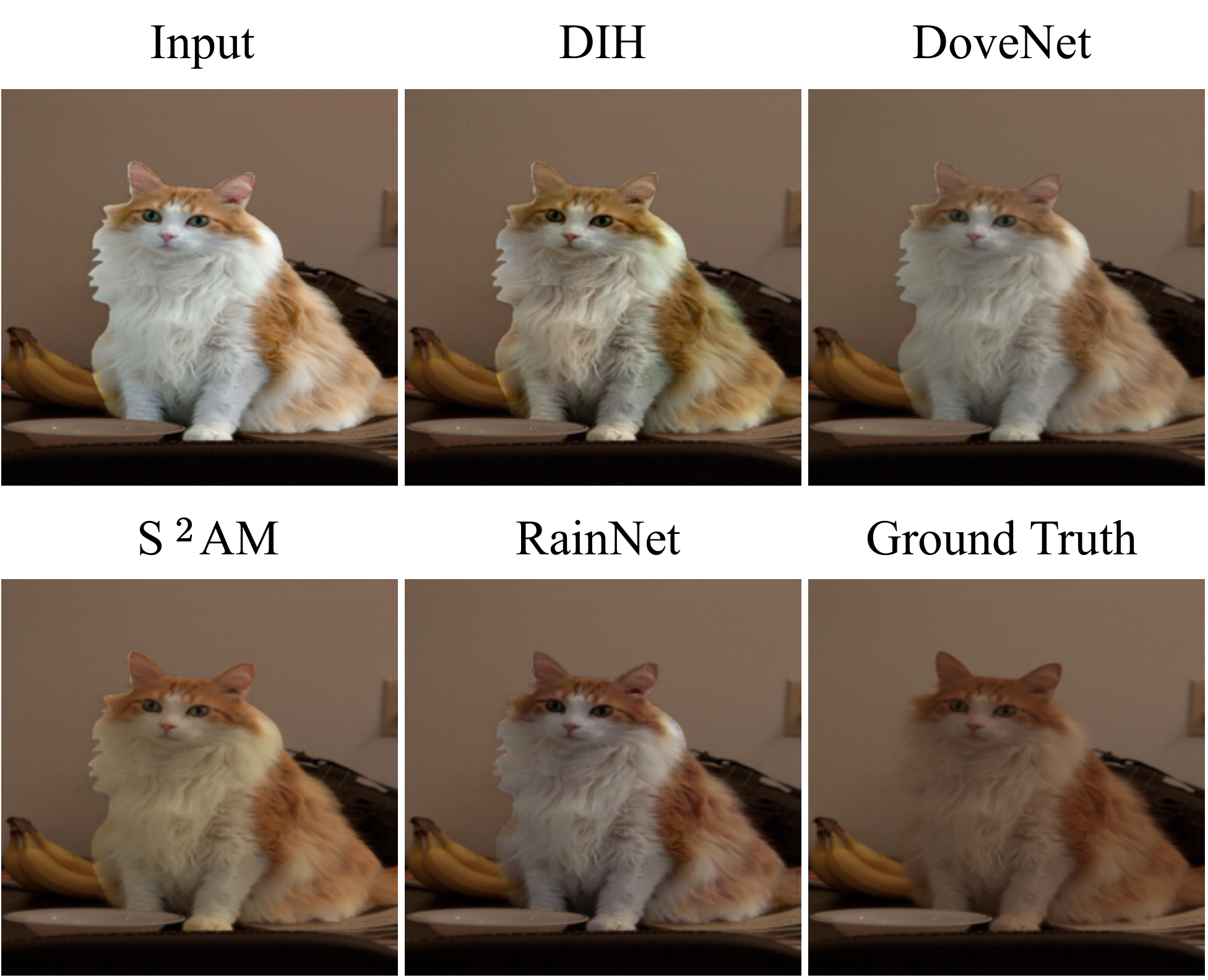}
\end{center}
   \caption{\textbf{Failure case.}. The proposed RainNet fails to harmonize the composite image with sharp foreground object and dim or blurry background image. }
   \label{fig:failure_cases}
\end{figure}

As has been refered in the main submission, the proposed RainNet fails to deal with the case of images with a blurred background with a sharp foreground object. Fig.~\ref{fig:failure_cases} shows an example. As can be found, S$^2$AM~\cite{cun2020improving} performs better than the proposed RainNet and other methods. However, these methods also fail to produce consistent boundary, introducing observable visual artifacts and deteriorating the visual quality. %Our future work should focus on this issue. 

\end{document}